\documentclass[10pt,twocolumn,letterpaper]{article}

\usepackage{iccv}
\usepackage{times}
\usepackage{epsfig}
\usepackage{graphicx}
\usepackage{amsmath}
\usepackage{amssymb}

\usepackage{subfigure}
\usepackage{multirow}
\usepackage[algoruled,vlined,linesnumbered]{algorithm2e}
\usepackage{rotating}
\usepackage{bbm}
\usepackage{bm}
\usepackage{color}
\usepackage{url}
\usepackage{threeparttable}
\usepackage{multirow}
\usepackage{rotating}
\usepackage{makecell}
\usepackage{diagbox}
\usepackage{cancel}

\usepackage[pagebackref=true,breaklinks=true,letterpaper=true,colorlinks,bookmarks=false]{hyperref}

\iccvfinalcopy % *** Uncomment this line for the final submission

\def\iccvPaperID{262} % *** Enter the ICCV Paper ID here

\ificcvfinal\fi
\begin{document}

%%%%%%%%% TITLE
\title{Attribute Recognition by Joint Recurrent Learning of Context and Correlation}

\author{\small Jingya Wang\\
	{\small Queen Mary University of London}\\
	{\tt\small jingya.wang@qmul.ac.uk}
	\and 
	\small Xiatian Zhu\\
	{\small Vision Semantics Ltd.}\\
	{\tt\small eddy@visionsemantics.com}
	\and
	\small Shaogang Gong\\
	{\small Queen Mary University of London}\\
	{\tt\small s.gong@qmul.ac.uk}
	\and
	\small Wei Li \\
	{\small Queen Mary University of London}\\
	{\tt\small wei.li@qmul.ac.uk}
}

\author{Jingya Wang$^1$ 
	\quad \quad \quad \quad Xiatian Zhu$^2$ 
	\quad \quad \quad \quad Shaogang Gong$^1$ 
	\quad \quad \quad \quad Wei Li$^1$ \\
Queen Mary University of London$^1$ \quad \quad \quad \quad %\\
% London E1 4NS, United Kingdom 
Vision Semantics Ltd.$^2$\\
{\tt\small \{jingya.wang, s.gong, wei.li\}@qmul.ac.uk} 
\quad \quad 
\tt\small eddy@visionsemantics.com
%\and
%Xiatian Zhu\\
%\and
%Shaogang Gong\\
%\and
%Wei Li\\
}

%\author{Jingya Wang, Xiatian Zhu, Shaogang Gong\\
%School of Electronic Engineering and Computer Science\\
%Queen Mary University of London\\
%London E1 4NS, United Kingdom\\
%\{jingya.wang, xiatian.zhu, s.gong\}@qmul.ac.uk
%}

\maketitle
%\thispagestyle{empty}

%%%%%%%%% ABSTRACT
%%%%%%%%% ABSTRACT
\begin{abstract} %{\bf \color{red} TODO:}
Recognising semantic pedestrian attributes in surveillance images 
	is a challenging task for computer vision,
	particularly when the imaging quality is poor with
	complex background clutter and uncontrolled viewing conditions,
	and the number of labelled training data is small.
	%
	%Existing methods either ignore the interaction between attributes 
	%or the informative context awareness, and thus suboptimal
	%in exploiting the limited amount of label data.
	% 
	In this work, we formulate a Joint Recurrent Learning (JRL)
        model for exploring attribute context and correlation in order
        to improve attribute recognition given small sized training
        data with poor quality images. 
        The JRL model learns jointly
        pedestrian attribute correlations in a pedestrian image and in
        particular their sequential ordering dependencies (latent
        high-order correlation) in an end-to-end encoder/decoder
        recurrent network.
%	modelling multiple context information and
%	the relationships among attributes and relevant spatial regions
%	Specifically, this is achieved by formulating a 
%	Joint Recurrent Learning (JRL) of attribute context and correlation
	%Recurrent Attribute Correlation Embedding
%	 network designed to discover the underlying high-order correlations between different attributes
%	and associate the spatial correspondence between image regions and attributes.
	%
%	As such, our JRL model has the potential of being robust against training data sparsity.
%	and remains a clear advantage over competitive methods
%	when the training data is limited.
	%
	We demonstrate the performance advantage and robustness of the JRL model over 
	a wide range of state-of-the-art deep models for pedestrian
        attribute recognition, multi-label image classification, and multi-person image annotation
	on two largest pedestrian attribute benchmarks PETA and RAP.
%	
%Pedestrian attributes recognition has been an hard topic in attribute recognition 
%as it is extremely fine-grained, the problem is more challenge in surveillance scenario considering the low quality of image, clustered background and view point variance for different attributes. Compare with traditional methods that treat attributes independently, some current research focus on jointly learning the attributes to model their relationships. However, both exiting methods only consider the pairwise relation among attributes. Inspired from human prediction process that describe images from more easy and holistic attribute to hard and specific ones, our model imitate same human cognition process, predict attributes sequentially, use previous prediction on easy attributes to help for hard ones.
%Moreover, existing methods loss the region relationship with attributes, e.g.shoes always appear on the bottom of the images and below the attribute trousers. In our proposed JRL  model, we jointly model the region-region context relation, attribute-attribute high order relation and attribute-region relation in a unified Encode-Decode attention-aware framework.  We demonstrate the advantages of our proposed model on public benchmark dataset.
\end{abstract}

\section{Introduction}

\begin{figure} %[th!]
	\centering
	\includegraphics[width=1\linewidth]{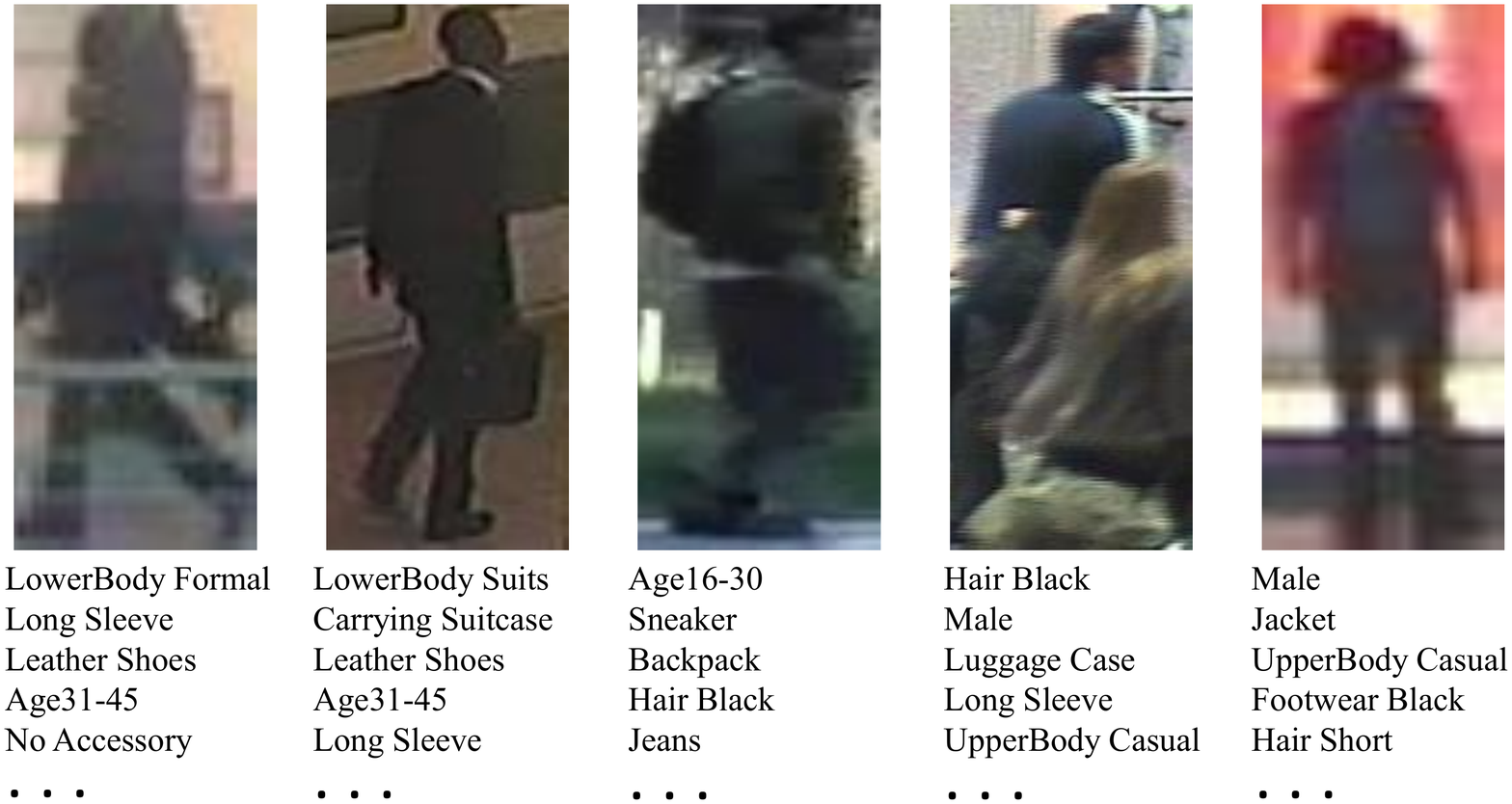}
	\vskip -0.8cm
	\caption{%\footnotesize
		Pedestrian attribute recognition in poor surveillance images 
		is inherently challenging
		due to ambiguous visual appearance from low resolution and 
		large variations in human pose and viewing conditions, e.g.
		illumination, background clutter, occlusion.
%		{\bf \color{cyan} TODO: Kiya to refine image selection for 
%			better reflecting the challenges in Intro.
%		Only need to show 3 hard attributes.}
		}
	\label{fig:problem}
	\vspace{-.3cm}
\end{figure}

% >> Attributes are important soft-biometrics or mid-level representation
Pedestrian attributes, e.g. age, gender, and hair style
are humanly searchable semantic descriptions and can be used as soft-biometrics in
visual surveillance, with applications in
person re-identification \cite{layne2012person,LiuEtAlECCV12,peng2016joint},
face verification \cite{Kumar09Attribute}, and
human identification \cite{reid2014soft}.
%
% >> The mertis of attributes
An advantage of attributes as semantic descriptions over low-level visual features is 
their robustness against viewpoint changes and viewing condition diversity.
% such as illumination and background clutters. % \cite{layne2012person,reid2014soft}.
%
% >> Challenge to recognise due to xx
However, it is inherently challenging to automatically recognise 
pedestrian attributes from real-world surveillance images % particularly
% those from real-world surveillance venues 
because: (1) The imaging quality is poor, in 
low resolution and subject to motion blur (Fig.~\ref{fig:problem});
%(2) people appearance can be significantly diverse
%in public scenes; and
(2) Attributes may undergo significant appearance changes and 
situate at different spatial locations in an image;
(3) Labelled attribute data from surveillance images are difficult to
collect and only available in small numbers.
These factors render learning a pedestrian attribute model very difficult.
Early attribute recognition methods  
mainly rely on hand-crafted features like colour and texture
\cite{layne2012person,LiuEtAlECCV12,jaha2014soft,DENG2014PAR}.
Recently, deep learning based attribute models have started to gain
attraction \cite{li2015multi,peng2017joint,dong2017multi,dong2017class}, due to deep model's
capacity for learning more expressive representations {\em when large scale
data is available}
\cite{krizhevsky2012imagenet,bengio2013representation,simonyan2014very}.
However, large scale training data is not available for
pedestrian attributes. The two largest pedestrian
attribute benchmark datasets PETA \cite{DENG2014PAR}
and RAP \cite{li2016richly} 
contain only $9,500$ and $33,268$ training images, % (image level annotation),
much smaller than the popular 
ILSVRC ($1.2$ million) \cite{ILSVRC15} 
and MS COCO ($165,482$) datasets \cite{lin2014microsoft}.
% which 
%have $1.2$ million and $165,482$  training images, respectively.
% provides $1.2$ million of images for model training alone 
% ImageNet: 1,281,167 (732-1300) 50,000 (50) 100,000 (100)
% MS COCO: 165,482 train, 81,208 val, and 81,434 test images. 328124 = 165482 + 81208 + 81434
%
%This data scarcity problem poses a huge challenge to typical deep models
%\cite{hoffman2013one}.
Deep learning of pedestrian attributes is further compounded by degraded fine-grained details due to poor
image quality, low resolution and complex appearance 
variations in surveillance scenes.

To address these difficulties, one idea is to discover the
interdependency and correlation among attributes 
\cite{chen2012describing,li2015multi,zhu2016multi,wang2016clustering,wang2017structure}, e.g. 
two attributes ``female'' and ``skirt'' 
are likely to co-occur in a person image. 
This correlation provides an inference constraint
complementary to visual appearance recognition.
Another idea is to explore visual context as an extra information source
to assist attribute recognition
\cite{li2016human,gkioxari2015contextual}. For instance, different people may
share similar attributes in the same scene, e.g. most skiers wear sun-glasses.
However, these two schemes are mostly studied independently by existing methods.
%
%The potential effects of modelling the latent inter-attribute correlation 
%have been demonstrated by graph model based methods 
%\cite{deng2015learning,chen2012describing,shi2015transferring}
%and recently by deep CNN models
%\cite{zhu2015multi,li2015multi,su2016deep,sudowe2015person,zhu2016multi}.
% 
%Consider that the visual appearance of surveillance images is inherently ambiguous,
%modelling attribute dependency and correlation is therefore essential \cite{xue2011correlative},
%e.g. female and long hair are very likely to co-exist on the same person.
%But, these existing methods are still rather limited,
%for example graph models rely on less expressive hand-crafted features and are not scalable to handle
%a large number of attributes, % (particularly graph models)
%whilst CNN models have little capability to model high-order relationships.

%characterised by 
%(1) jointly learning multiple pedestrian attributes,
%(2) simultaneously modelling inter-attributes correlation at varying orders,
%(3) jointly embedding local and global context information,
%(4) automatically identifying attribute relevant regions. 

% >> Our method
In this work, we explore {\em both} modelling inter-person image context and learning
intra-person attribute correlation in a unified framework. %, which is still significantly under-studied.
To this end, we formulate a Joint Recurrent Learning ({\bf JRL}) 
of attribute correlation and context.
We introduce a novel Recurrent Neural Network (RNN) encoder-decoder architecture
specifically designed for {\em sequential} pedestrian attribute prediction
jointly guided by both {\em intra-person attribute} and {\em
  inter-person similarity} context awareness.
%The proposed model is capable of 
%not only learning multiple pedestrian attributes jointly,
%and modelling low-/high-order inter-attribute correlation simultaneously,
%but also mining the localised person spatial (intra-image) context and
%incorporating the visually similar exemplar (inter-image) context.
%
This RNN based model explores explicitly a {\em sequential} prediction
constraint that differs significantly from
the existing CNN based {\em concurrent} prediction policy
\cite{li2016human,gkioxari2015contextual,li2015multi}. We argue that this approach
enables us to exploit more latent and richer higher-order
dependency among attributes, therefore better mitigating the
small training data challenge.
Our approach is motivated by natural language sentence prediction
which models inter-word relations \cite{vinyals2015show,liu2016semantic}.
% \kiya{and easy-to-hard curriculum learning method \cite{pentina2015curriculum}.}
Importantly, two information sources (intra-person attribute correlations and
inter-person image similarities) 
are {\em simultaneously} modelled to learn person-centric inter-region
correlation to compensate poor (or missing) visual details.
This provides the model with a more robust embedding given poor
surveillance images and learns more accurate intra-person attribute correlations.
Crucially, we do not assume people in the same scene share
common attributes \cite{li2016human}, nor assuming 
person body-part detection \cite{gkioxari2015contextual}. 
Because people appearances in surveillance scenes are without a common
theme, and person body-part detection in low
resolution images under poor lighting is inconsistent, resulting in many poor
detections. 

More specifically, our approach considers
%the recurrent neural networks (RNN) Encoder-Decoder network architecture
the {\em sequence-to-sequence mapping} framework
(with a paired encoder and decoder)
% and either is an RNN network
\cite{cho2014properties,sutskever2014sequence,cho2015describing}.
%our model consists of two paired recurrent neural networks (RNN):
%one is encoder and the other decoder.
To explore a sequence prediction model,
we convert any given person image
into a region sequence
(Fig.~\ref{fig:pipeline}({b})) and a set of attributes into an ordered list
(Fig.~\ref{fig:pipeline}({c})). 
An {\em encoder} maps a {\em fixed-length} image region sequence 
into a continuous feature vector. 
The recurrent step is to encode sequentially localised person spatial context 
and to propagate inter-region contextual information. We call this
{\em intra-person attribute context} modelling. 
Moreover, we also incorporate {\em inter-person similarity context} (Fig.~\ref{fig:pipeline}({a})).
That is, 
we identify % $k$ ($k=2$ in our experiments) 
visually similar exemplar images in the training set,
encode them so to be combined with the encoded image by
similarity max-pooling.
This fused encoding feature representation is used to initialise a decoder.
%for regularising a better initial state.
%
The {\em decoder} transforms the image feature vector from the encoder
to a {\em variable-length} attribute sequence as
output.
%(\sgg{inconsistent - we said elsewhere that the output is a fixed
%  length attribute sequence. Check and correct accordingly})
This joint sequence-to-sequence encoding and decoding
process enables a low- to high-order attribute correlation learning in
a unified model.
%
%To mitigate the effect of noisy visual appearance, 
%we exploit the exemplar context from visually similar training images
%to more accurately regularise the decoder initial states. 
%By such sequential attribute modelling, 
%As such, our method allows to not only encode the spatial local dependency and 
%propagate relevant contextual information through the encoder network,
%
%but also %jointly learn multiple pedestrian attributes
%capture high-order correlation among attributes
%in a joint manner.
% since 
%our whole model is optimised to do so in the training stage.
%The two RNN networks are trained jointly to maximise the conditional probability of 
%the attribute sequence given the ground-truth attribute sequence. 
% >> 
As attributes are weakly-labelled at the image level without
fine-grained localisation, 
we further exploit a data-driven
attention mechanism \cite{bahdanau2014neural} to %automatically
identify attribute sensitive image regions and to guide the decoder to
those image regions for feature extraction.

The {\bf contributions} of this work are:
{\bf (1)} We propose a Joint Recurrent Learning (JRL) approach to 
pedestrian attribute correlation and context learning in a unified model.
This is in contrast to existing methods that
separate the two learning problems thus suboptimal
\cite{li2016human,gkioxari2015contextual,li2015multi,sudowe2015person}.
{\bf (2)} We formulate a novel end-to-end encoder-decoder architecture
capable of jointly learning image level context and
attribute level sequential correlation.
%{\em Recurrent Attribute Correlation Embedding}
%(JRL) method specialised at predicting attributes sequentially
%by jointly considering the spatial context relationship,
%correlation between attributes, and the correspondence between
%attributes and image regions in a unified network.
To our best knowledge, this is the first attempt of formulating 
{\em pedestrian attribute recognition as a sequential prediction problem}
designed to cope with poor imagery data with missing details.
{\bf (3)} We provide extensive comparative evaluations
on the two largest pedestrian attribute benchmarks (PETA \cite{DENG2014PAR} and 
RAP \cite{li2016richly}) against 7 contemporary models including
5 pedestrian attribute models (SVM \cite{layne2014attributes}, MRFr2
\cite{deng2015learning}, ACN \cite{sudowe2015person}, DeepSAR and DeepMAR \cite{li2015multi}), 
a multi-label image classification model
(Semantically Regularised CNN-RNN \cite{liu2016semantic}), %, RIA \cite{jin2016annotation})
and a multi-person image annotation model (Contextual CNN-RNN \cite{li2016sequential}).
The proposed JRL model yields superior performance compared to these methods.
\section{Related Work}

\noindent {\bf Pedestrian Attribute Recognition.}
Semantic pedestrian attributes have been extensively exploited for
person identification \cite{jaha2014soft} 
and re-identification \cite{layne2012person,su2016deep,peng2016joint}.
Earlier methods typically model multiple attributes independently
and train a separate classifier for each attribute (e.g. SVM or AdaBoost)
based on hand-crafted features such as colour and texture histograms
\cite{layne2012person,zhu2013pedestrian,layne2014attributes,DENG2014PAR}.
% Clearly, this strategy is unscalable to a large number of attributes,
%, if the number of attributes is small; 
%these methods ignore the interactions between different attributes.
% 
%\cite{layne2012person,layne2014attributes,jaha2014soft}:
%each attribute classifier is trained by using a support vector machine (SVM)
%
%\cite{zhu2013pedestrian}:
%the gentle AdaBoost algorithm is applied to train attribute classifier independently
%
Inter-attribute correlation was 
considered as complementary information to compensate noisy visual appearance
for improving prediction performance, e.g. graph model based methods allow to
capture attribute co-occurrence likelihoods by 
using conditional random field or Markov random field 
to estimate the final joint label probability \cite{deng2015learning,chen2012describing,shi2015transferring}.
However, these methods are prohibitively expensive to compute
when dealing with a large set of attributes,
due to the huge number of model parameters on pairwise relations.
Deep CNN models have been exploited for joint multi-attribute
feature and classifier learning \cite{zhu2015multi,li2015multi,sudowe2015person,zhu2016multi,li2016human,dong2017class},
%\kiya{(\cite{li2016human} address Q11 from R2: “MissingY.Lietal. ECCV2016}
and shown to benefit from learning attribute co-occurrence dependency.
However, they do not explore modelling high-order attribute sequential correlations.
Other schemes also exploited contextual information
\cite{gkioxari2015contextual,li2016human}, but making too strong
assumptions about image qualities to be applicable to surveillance data.
Attribute correlation and contexting are often treated
independently by existing methods. This work aims to explore
jointly their complementary benefits in improving attribute
recognition when only small sized and poor quality training data is
available.

\vspace{0.1cm}
%====================
\noindent {\bf Multi-Label Image Classification.}
Pedestrian attribute recognition is a Multi-Label Image Classification (MLIC) problem
\cite{makadia2008new,guillaumin2009tagprop}.
%,.
Sequential multi-label prediction has been explored before \cite{wang2016cnn,liu2016semantic}. %,jin2016annotation}.
These methods are based on a CNN-RNN model design, whilst our JRL model has a CNN-RNN-RNN architecture.
Crucially, these existing MLIC models assume (1) the availability of
large scale labelled training data (2) with sufficiently good image
quality, e.g. $165,482$ carefully selected Flickr photo images in the MS COCO dataset \cite{lin2014microsoft}.
Both assumptions are invalid for pedestrian attribute recognition in
surveillance images. A very recent multi-person image annotation method advances this
sequential MLIC paradigm by incorporating additional inter-person social relations
and scene context \cite{li2016sequential}.
This method exploits specifically context among family members and friend-centric photo images 
of high-resolution, but not scalable to open-world surveillance scenes
of poor image data. Moreover, strong attribute-level labels are
required \cite{li2016sequential}, %(\sgg{what's their training data size})
whilst pedestrian attributes are mostly weakly-labelled at the image level.
In contrast, the
proposed JLR model is designed specially to combat the challenges of
attribute recognition in low-resolution poor quality images with
weakly-labelled data in small training data size. 
%through learning inter-region topological correlation (i.e. intra-person spatial context)
%and incorporating visually similar exemplars (i.e. inter-person compensation context)
%in a joint fashion.
Learning image region sequence correlation has been exploited 
for face recognition \cite{samaria1993face} and 
person re-identification \cite{varior2016siamese}.
Their problem settings are different from this work:
here we aim to exploit image level context for enhancing sequential 
attribute correlation learning in a multi-label classification
setting, whist both \cite{samaria1993face} and
\cite{varior2016siamese} consider a single-label image classification problem.
\section{Joint Recurrent Learning of Attributes}

%% Why correlation
% 1) Challenges of recognising attributes
% 2) Data scarcity
% 3) Image appearance ambiguouty

%==============================
%\subsection{Problem Definition}
To establish a deep model for recognising pedestrian attributes in
inherently ambiguous surveillance images, we assume $n$ labelled training images
$\mathcal{I}_\text{tr} = \{\bm{I}_i\}_{i=1}^{n}$ available, 
with the attribute labels as 
$\mathcal{A}_\text{tr} = \{\bm{a}_i\}_{i=1}^{n}$.
Each image-level label annotation $\bm{a}_i = [a_{(i,1)}, \dots, a_{(i,n_\text{attr})}]$ 
is a binary vector, defined over $n_\text{attr}$ 
pre-defined attributes with $0$ and $1$ indicating the absence and presence 
of the corresponding attribute, respectively.
Intrinsically, this is a {\em multi-label} recognition problem 
since the $n_\text{attr}$ pedestrian attribute categories may co-exist in a single image.
It is necessary to learn {\em attribute sequential correlations
  (high-order) in similar imagery context} in order to overcome the
limitations in training data due to poor image quality, weak
labelling, and small training size. 
%
%
%
%However, it is non-trivial to model the intrinsically complex
%inter-attribute relationships 
%given the great diversity in people appearance at public scenes
%captured by poor images and sparse weakly labelled attribute annotations.
%That is, there may exist different orders of dependency and 
%it is challenging to jointly learn them in a unified model.
%The learning challenge can be further elevated
%if the available training data is small. 
%%
%On the other hand, labels are provided in image level
%but many attributes are {\em localised} to image regions
%with unknown and complex spatial correspondence patterns involved
%due to the large variations in pose and viewing conditions.
%We call this {\em weak label}.
%
To that end, we formulate a Joint Recurrent Learning (JRL) of both
attribute context and correlation in an end-to-end sequential deep model.
% correlation driven pedestrian attribute recognition model,
%as detailed in follows.

\begin{figure*}%[th!]
	\centering
	\includegraphics[width=0.85\linewidth]{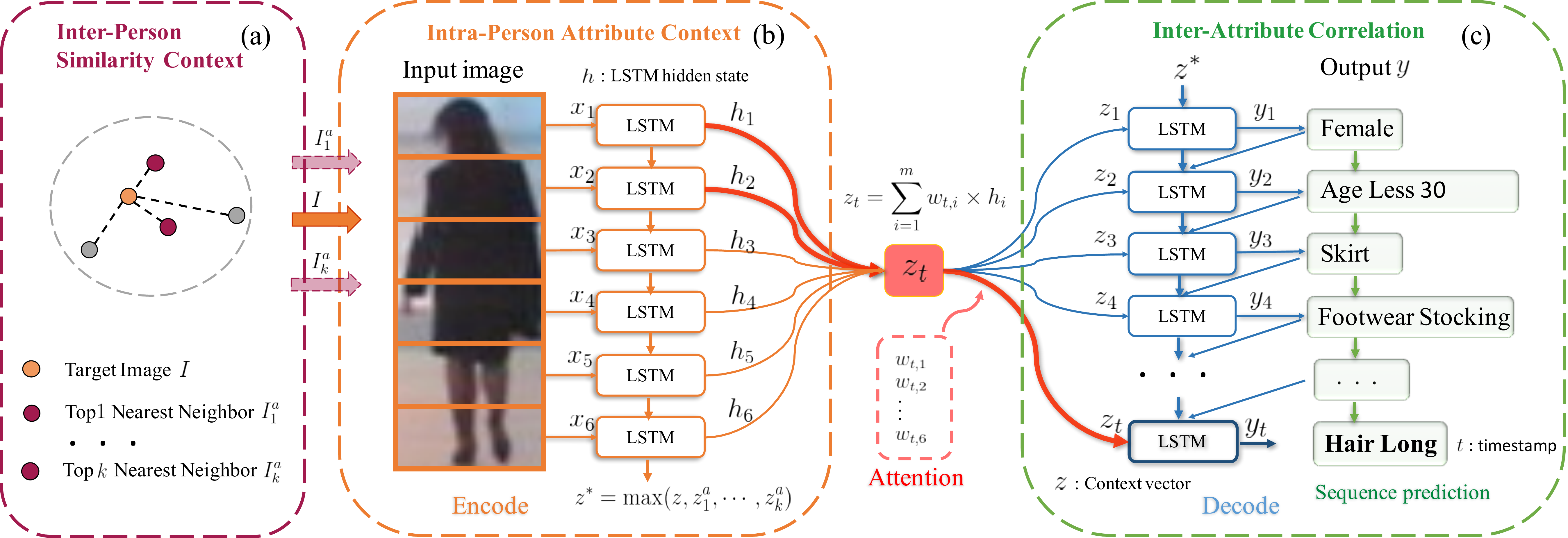}
	%\vskip -.1cm
	\caption{%\footnotesize
		An overview of the proposed Joint Recurrent Learning (JRL) of attribute context and correlation.
	%{\bf \color{cyan} TODO: Kiya to revise.}	
}
	\label{fig:pipeline}
	\vspace{-.5cm}
\end{figure*}

\subsection{Network Architecture Design}
\label{sec:JRL}

% \noindent {\bf Network Architecture.}
An overview of the proposed JRL architecture is depicted in 
Fig. \ref{fig:pipeline}.
We consider the RNN encoder-decoder framework
as our base model
%, where two RNNs 
%(i.e., one as encoder and the other as decoder) are contained
%
%This is because this architecture is 
because of its powerful capability in learning sequence data
% and well suited to 
and modelling the translation between 
different types of sequences
\cite{sutskever2014sequence,cho2014learning,cho2014properties}.
Specifically, RNN is a neural network
consisting an internal hidden state $\bm{h} \in \mathbb{R}^d$
% and an optional output
and operating on a
% specialised at handling 
variable-length input sequence
$X = (\bm{x}_1, \dots, \bm{x}_t, \dots)$.
%and corresponding arbitrary-length output sequences 
%$Y = (\bm{y}_1, \dots, \bm{y}_t, \dots)$
%by maintaining an internal hidden state $\bm{h}_t$ over time $t$.
At each time step $t$,
the RNN takes sequentially an element $\bm{x}_t$ of $X$
and then updates its hidden state $\bm{h}_t$ as
\begin{equation}
\bm{h}_t = \phi_\theta (\bm{h}_{t-1}, \bm{x}_t)
\label{eqn:rnn}
\end{equation}
where $\phi_\theta$ denotes the non-linear activation function parameterised
by $\theta$.
%
%
%{\em LSTM.} 
To capture the long range dependency of attribute-attribute, region-region,
and attribute-region, we adopt the LSTM \cite{hochreiter1997long} 
as recurrent neuron for both encoder and decoder RNN.
Also, LSTM is effective to handle the common gradient 
vanishing and exploding problems 
in training RNN \cite{pascanu2013difficulty}.
Particularly, at each time step $t$, the LSTM updates using
the input $\bm{x}_{t}$ and the LSTM previous status $\bm{h}_{t-1} \in \mathbb{R}^d$, 
and $\bm{c}_{t-1} \in \mathbb{R}^d$ as:
\begin{align} %\small
\begin{split}
\bm{f}_{t} & = \mbox{sigmoid} (\bm{W}_{fx}\bm{x}_{t}+\bm{W}_{fh}\bm{h}_{t-1}+\bm{b}_{f}) \\
\bm{i}_{t} & = \mbox{sigmoid} (\bm{W}_{ix}\bm{x}_{t}+\bm{W}_{ih}\bm{h}_{t-1}+\bm{b}_{i}) \\
\bm{o}_{t} & = \mbox{sigmoid} (\bm{W}_{ox}\bm{x}_{t}+\bm{W}_{oh}\bm{h}_{t-1}+\bm{b}_{o}) \\
\bm{g}_t & = \mbox{tanh}(\bm{W}_{gx}\bm{x}_{t} + \bm{W}_{gh}\bm{h}_{t-1} + \bm{b}_{g}) \\
\bm{c}_{t} & = \bm{f}_{t} \odot \bm{c}_{t-1} + \bm{i}_{t} \odot \bm{g}_t \\ % \mbox{tanh}(\bm{W}_{gx}\bm{x}_{t} + \bm{W}_{gh}\bm{h}_{t-1} + \bm{b}_{c}) \\
\bm{h}_{t} & = \bm{o}_{t} \odot \mbox{tanh}(\bm{c}_{t})
\end{split}
\label{enq:lstm}
\end{align}
where 
$\mbox{sigmoid}(\cdot)$ refers to the logistic sigmoid function, 
%(the rectified linear units (ReLU) \cite{dahl2013improving}
%in our implementation),
$\mbox{tanh}(\cdot)$ the hyperbolic tangent function,
the operator $\odot$ the element-wise vector product.
The LSTM contains four multiplicative gating units:
% 
% Also, the LSTM includes 
forget gate $\bm{f} \in \mathbb{R}^d$,
input gate $\bm{i} \in \mathbb{R}^d$,
output gate $\bm{o} \in \mathbb{R}^d$,
input modulation gate $\bm{g} \in \mathbb{R}^d$,
with matrix $\bm{W}$s and vector $\bm{b}$s the corresponding to-be-learned parameters.
The memory cell $\bm{c}_t$ depends on 
(1) the previous memory cell $\bm{c}_{t-1}$, modulated by $\bm{f}_t$, and 
(2) the input modulation gate, %a function of $\bm{x}_t$ and $\bm{h}_{t-1}$, 
modulated by $\bm{i}_t$.
As such, the LSTM learns to
forget its previous memory and exploit its current input
selectively.
Similarly, the output gate $\bm{o}$ learns how to transfer the memory cell 
$\bm{c}_t$ to the hidden state $\bm{h}_t$.
%
%In particular, LSTM contains three multiplicative units: 
%(1) a forget gate $\bm{f} \in \mathbb{R}^d$ that controls whether to forget the current state, 
%(2) an input gate $\bm{i} \in \mathbb{R}^d$ that indicates if it should read the input, 
%(3) an output gate $\bm{o} \in \mathbb{R}^d$ that governs whether to output the state.
Collectively, these gates learn to effectively modulate
the behaviour of input signal propagation
% effectively modulate and propagate the input signal
through the recurrent hidden states
for helping capture complex and long-term dynamics/dependency in sequence data.
%e.g., by back-propagation through time \cite{werbos1990backpropagation}.

\vspace{0.1cm}
\noindent {\bf (I) Intra-Person Attribute Context.} 
We model the intra-person attribute context 
within each person image $\bm{I}$ by the encoder LSTM.
This is achieved by mapping recurrently each input 
%(\sgg{not just
%  target image, should be every image, including both training and test image?})
image into a fixed-length feature vector (Fig.~\ref{fig:pipeline} (b)).
%The purpose is to encode intra-person topological context 
%information of the target person into the representation.
Specifically, for allowing sequential modelling of the person image $\bm{I}$, 
we first divide it into 
$m$ (empirically set $m=6$ in our experiment) 
horizontal strip regions and form a region sequence
${S} = (\bm{s}_1,\dots, \bm{s}_m)$ in top-bottom order.
Then the encoder reads each image region sequentially, 
and the hidden state $\bm{h}^\text{en}$ of the encoder LSTM
updates according to Eqn. \eqref{enq:lstm}.
Once reaching the end of this region sequence,
the hidden state $\bm{h}_m^\text{en}$ of the encoder can
be considered as the summary representation $\bm{z} = \bm{h}_m^\text{en}$
of the entire sequence or the person image.
We call $\bm{z}$ as {\em context vector}.
Importantly, this allows to selectively extract and encode 
the spatial dependency between different body parts whilst
also propagate relevant localised topological contextual 
information through the recurrent network,
thanks to the capability of LSTM in 
modelling the long short term relationships between sequence elements.

\vspace{0.1cm}
\noindent {\bf (II) Inter-Person Similarity Context.}
To compensate appearance ambiguity and poor image quality in a target image,
we explore auxiliary information from visually similar exemplar training images
to provide a inter-person similarity context constraint.
%This modelling is performed by the encoder LSTM.
Specifically, we search top-$k$ exemplars $\{\bm{I}_i^a\}_{i=1}^k$
that are visually similar to the image $\bm{I}$ 
from the training pool. % by using the deep CNN feature and the generic L2 metric.
For each exemplar $\bm{I}_i^a$, we compute its 
own context vector %image feature representation 
$\bm{z}_i^a$
using the same encoding process as that for the image $\bm{I}$.
Then, we ensemble all the context vector representations of auxiliary
images as the inter-person context to $\bm{z}$ as follows:
\begin{equation}
\bm{z}^* = \max(\bm{z}, \bm{z}_1^a, \cdots, \bm{z}_k^a)
\label{eqn:max_fusion}
\end{equation}
where $\max(\cdot)$ defines the element-wise maximum operation over all
input feature vectors of both the input image and top-$k$ exemplars.
While the averaging based ensemble may be more conservative and reducing the
likelihood of introducing additional noisy information, we found empirically
that maximum-pooling based ensemble is more effective.
The rational for this inter-person similarity context compensation is that
missing or corrupted local information in the input image
cannot be easily recovered in the decoding process whilst 
newly introduced localised noise can be largely suppressed by 
	optimising the decoder.
%(\sgg{i rephrased
%  this - check carefully})

\vspace{0.1cm}
\noindent {\em Image Representation and Similarity Search.}
As input to the LSTM encoder, we utilise a deep CNN initialised by
ImageNet (e.g. the
AlexNet \cite{krizhevsky2012imagenet}), then fine-tune the CNN on the
pedestrian attribute training data to better represent pedestrian
images by its deep feature vectors. Specifically, for a given person image, 
we decompose the activations of the $5^\text{th}$ convolutional layer
into $m$ horizontal regions, 
each of which is pooled 
into a vector by directly concatenating all dimensions.
Moreover, we use the FC$_7$ layer's output as the feature space for 
top-$k$ exemplar similarity search using L2 distance metric. 
%(each location in the feature map corresponds to a regular region in the input image), 
%\hl{Then, we add a linear projection function to project region based representation into $l$ dimensional space ($l$ is the dimension of LSTM hidden state)}.

\vspace{0.1cm}
%\noindent {\bf (III) Decoder LSTM.}
\noindent {\bf (III) Inter-Attribute Correlation.}
We construct a decoder LSTM to model 
the latent high-order attribute correlation subject to jointly
learning a multi-attribute prediction classifier.
%	Specifically, we start with initialising the hidden state of the decoder 
%	with the improved context vector as $\bm{h}^\text{de}_{1} = \bm{z}^*$
%	the above attribute decoding process by setting
%$\bm{h}^\text{de}_{1} = \bm{z}$, and 
%$\bm{y}_{t-1} = \bm{0}$.
%
Specifically,
given the encoded context vectors $\bm{z}$ and $\bm{z}^*$,
%from the encoded ensemble context $\bm{z}$ 
%(\sgg{should
%  this be $\bm{z}^*$? $\bm{z}$ is the target not the context})
%(the main interface between the encoder and decoder),  
the decoder LSTM aims to model a sequential recurrent attribute
correlation within both
{\em intra-person attribute context} ($\bm{z}$) and 
{\em inter-person similarity context} ($\bm{z}^*$)
and to generate its variable-length output as a predicted sequence of attributes $\bm{y}_t$ over time $t$
(Fig.~\ref{fig:pipeline}(c)).
This is desired since the co-occurring attribute number varies among individual images.
An attribute label sequence of a person image is generated 
from a fixed order list of all attributes 
(Sec.~\ref{sec:model_train_test}).
We initialise the decoder hidden state $\bm{h}^\text{de}_{1}$ 
with the improved encoder context vector: $\bm{h}^\text{de}_{1} = \bm{z}^*$.
This is to incorporate the {\em inter-person similarity context} into the decoding process.
Compared to the encoder counterpart,
$\bm{h}^\text{de}_{t}$ and $\bm{y}_t$ are additionally conditioned on
the previous output $\bm{y}_{t-1}$ 
(initialised $\bm{y}_{0} = \bm{0}$, i.e. the ``start'' token).
%$\bm{y}_{0} = \bm{0}$
In essence, it is this sequential recurrent feedback connection that 
enables our model to mine the varying high-orders of 
attribute-attribute dependency -- 
longer prediction, higher-order attribute correlation modelled.
Formally, rather than by Eqn. \eqref{eqn:rnn}$,
\bm{h}^\text{de}_{t}$ is updated via:
\begin{equation}
 \bm{h}_{t}^\text{de} = \phi_{\theta}(\bm{h}_{t-1}^\text{de}, \bm{y}_{t-1}, \bm{z}).
\label{eqn:hidden_decoder}
\end{equation}
In case of LSTM, the particular update formulation is:
\begin{align} \small
\begin{split}
\bm{f}_{t} & = \mbox{sigmoid} (\bm{W}_{fz}\bm{z}+\bm{W}_{fh}\bm{h}_{t-1}^\text{de}+\bm{W}_{fy}\bm{y}_{t-1}+\bm{b}_{f}) \\
\bm{i}_{t} & = \mbox{sigmoid} (\bm{W}_{iz}\bm{z}+\bm{W}_{ih}\bm{h}_{t-1}^\text{de}+\bm{W}_{iy}\bm{y}_{t-1}+\bm{b}_{i}) \\
\bm{o}_{t} & = \mbox{sigmoid} (\bm{W}_{oz}\bm{z}+\bm{W}_{oh}\bm{h}_{t-1}^\text{de}+\bm{W}_{oy}\bm{y}_{t-1}+\bm{b}_{o}) \\
\bm{g}_t & = \mbox{tanh}(\bm{W}_{gz}\bm{z} + \bm{W}_{gh}\bm{h}_{t-1}^\text{de}+\bm{W}_{gy}\bm{y}_{t-1} + \bm{b}_{g}) \\
\bm{c}_{t} & = \bm{f}_{t} \odot \bm{c}_{t-1} + \bm{i}_{t} \odot \bm{g}_t \\ % \mbox{tanh}(\bm{W}_{gx}\bm{x}_{t} + \bm{W}_{gh}\bm{h}_{t-1}+\bm{W}_{cy}\bm{y}_{t-1} + \bm{b}_{c}) \\
\bm{h}_{t} & = \bm{o}_{t} \odot \mbox{tanh}(\bm{c}_{t})
\end{split}
\label{enq:lstm_decoder}
\end{align}
This is similar to Eqn. \eqref{enq:lstm} except the extra dependence on the previous prediction
$\bm{y}_{t-1}$ and corresponding parameters\footnote{We do not
  use the encoded $\bm{z}^*$ with inter-person similarity context 
  as the
  decoder input to avoid possible divergence from the exemplar
  images.}.
To predict the attribute, we first compute the conditional probability
over all attributes and a ``stop'' signal as: % $\bm{y}_{t}$, 
\begin{align}
\begin{split}
\bm{p}(\{{y}_{t,i} = 1\}_{i=1}^{n_\text{attr}+1}) 
%| \bm{y}_{t-1},\bm{y}_{t-2}...,\bm{y}{_{1}},\bm{z}
& = \phi_y(\bm{h}_{t-1}^\text{de}, \bm{y}_{t-1}, \bm{z}) \\ %\nonumber
& = \bm{W}_y \bm{o}_t + \bm{b}_y
\end{split}
\label{eqn:attr_prob}
\end{align}
where $\bm{W}_y \in \mathbb{R}^{(n_\text{attr}+1) \times d}$ 
and $\bm{b}_y \in \mathbb{R}^{(n_\text{attr}+1)}$ are model parameters,
and $\bm{o}_t \in \mathbb{R}^{d}$ can be obtained by Eqn. \eqref{enq:lstm_decoder}.
%it is actually a 
%conditional probability computed as:
%
Then, we predict the current attribute $\bm{y}_t$ as:
% perform attribute $\bm{y}_1$ prediction with
\begin{equation}
i^* = \mbox{argmax}_{i \in [1,\dots,n_\text{attr}+1]} ({{y}_{t,i}}),
\label{eqn:attr_pred}
\end{equation}
i.e. the $i^*$-th bit of $\bm{y}_t$ is $1$ whilst all the others are $0$.
%
%Where $s_{i}^{t}$ is the score of attribute index $i$ and $V$ 
%is the total attribute size plus STOP signal.
%Note that we start the above attribute decoding process by setting
%$\bm{h}^\text{de}_{1} = \bm{z}$, and 
%$\bm{y}_{t-1} = \bm{0}$. 

\vspace{0.1cm}
\noindent {\em Recurrent Attribute Attention.}
% Why attend to regions
%1) \cite{cho2014properties}: 
%attribute patterns can be vary in a complex degree and such information can be largely discarded when summarising an person image into a single fix-length feature vector
%due to the limited representation capacity of the fixed-dimensional context vector.
%
Appearance attribute patterns in real-world person images can vary complexly and significantly. 
By summarising a person image into a single fixed-length context vector $\bm{z}$
with the encoder,
a large amount of semantic information, (e.g. spatial distribution)
may be not well encoded due to the limited representation capacity of context vector \cite{cho2014properties}.
To overcome this limitation, 
we propose to improve our JRL model by incorporating the attention mechanism \cite{bahdanau2014neural,cho2015describing}
so as to automatically identify and focus on the most relevant parts 
of the input region sequence when predicting the current attribute
to improve the correlation modelling and finally the prediction performance.
This is essentially an explicit sequence-to-sequence alignment mechanism.
We achieve this by imposing a structure into the encoder output
and then reformulating the attribute decoding algorithm accordingly.

Specifically, we first allow the encoder to output 
a structured representation, a set of summary vectors as: 
\begin{equation}
H^\text{en} = (\bm{h}_{1}^\text{en}, \dots, \bm{h}_{i}^\text{en},....,\bm{h}_{m}^\text{en}), \quad \bm{h}_{i}^\text{en} \in \mathbb{R}^d
\label{eqn:encoder_output_struct}
\end{equation}
for an input image region sequence ${S} = (\bm{s}_1,\dots, \bm{s}_m)$
with $m$ the number of all time steps or the input region sequence length.
Clearly, $\bm{h}_{i}^\text{en}$ represents the context representation 
of the $i$-th (top-down order) spatial region of the input image.
The aim of our attribute attention is to 
identify an optimised weighting distribution 
$\bm{w}_t = (w_{t,1}, \cdots, w_{t,i}, \cdots, w_{t,m})$ over 
$(\bm{h}_{1}^\text{en}, \cdots, \bm{h}_{i}^\text{en}, \cdots ,\bm{h}_{m}^\text{en})$
at each time step $t$ when the decoder is predicting the attribute, as:
\begin{equation}
w_{t,i} = \frac{\exp(\alpha_{t,i})}{\sum_{i = 1}^{m}
  \exp(\alpha_{t,i})}, \mbox{with}\,\, \alpha_{t,i} = \phi_\text{att} (\bm{h}_{t-1}^\text{de}, \bm{h}^\text{en}_{i})
\label{eqn:attention_score_nrom}
\end{equation}
where $\phi_\text{att}$ defines the attention function
realised with a feed forward neural network in our 
approach as in \cite{vinyals2015grammar}.
%in our encode-decode attribute prediction framework following by \cite{bahdanau2014neural}. We first allow encoder return a structured representation of $k$ input strip: $ C={c_{1},c_{2},....,c_{k}} $, and current hidden state for decoder LSTM $si$ for time $i$ is :
%
Once obtaining the attention weighting score $\bm{w}_t$, we utilise it
to compute the step-wise context representation $\bm{z}_t$ by
\begin{equation}
\bm{z}_{t} = \sum_{i = 1}^{m} w_{t,i} \times \bm{h}_{i}^\text{en}
\label{eqn:summary_feat_attention}
\end{equation}
The final prediction $\bm{y}_t$ can be obtained similarly by
computing $\bm{h}_t^\text{de}$ with Eqn. \eqref{eqn:hidden_decoder},
and applying Eqns. \eqref{eqn:attr_prob} and \eqref{eqn:attr_pred}.

Note that the context representation $\bm{z}_t$ utilised 
in attention-aware attribute decoding
is varying over time $t$ due to the difference in the spatial attention 
distribution $\bm{w}_t$.
In contrast, in case of no attention,  %(Eqn. \eqref{eqn:hidden_decoder})
$\bm{z}$ is constant during the whole decoding process.
Implicitly, the current $\bm{w}_t$ is conditioned on $\bm{w}_{t-1}$ 
(the attention used at time $t-1$)
due to the dependence of $\bm{h}_{t-1}$ on $\bm{z}_{t-1}$ 
(Eqn. \eqref{eqn:hidden_decoder}) and 
of $\bm{z}_{t-1}$ on $\bm{w}_{t-1}$ (Eqn. \eqref{eqn:summary_feat_attention}), 
therefore allowing to optimise
the underlying correlation in per-step attention selection for 
sequential attribute decoding during model training.

% ======= Attribute embedding
\vspace{0.1cm}
\noindent {\em Attribute Embedding.}
To incorporate the previous attribute prediction as recurrent feedback on
the next prediction, we need a proper attribute representation.
One common way is the simple one-hot vector.
Alternatively, word embedding \cite{mikolov2013efficient} has been shown as
a favourable text representation by learning
a lookup table optimised for specific annotation error 
and thus more semantically meaningful.
Therefore, we adopt the latter scheme in our attribute decoder.

\subsection{Model Training and Inference}
\label{sec:model_train_test}
%{\color{red} TODO: add some words.}

%\noindent \eddy{\bf Attribute Order Ensemble.}
To train the JRL model, we need to determine attribute selection order.
	However, pedestrian attributes are naturally unordered without a fixed ordering,
	similar to generic multi-label image classification \cite{liu2016semantic,wang2016cnn} and dissimilar to
	image caption \cite{vinyals2015show}. 
	To address this problem, one can either randomly fix an order
        \cite{li2016sequential} or define some occurrence frequency
        based orders, e.g. rare first (rarer attributes are placed earlier so that
	they can be promoted) \cite{liu2016semantic}, or frequent first 
	(more frequent attributes appear earlier with the intuition
	that easier ones can be predicted before harder ones) \cite{wang2016cnn}.
	In our model training, we explore the ensemble idea \cite{read2011classifier}
	to incorporate the complementary benefits of all these different orders
	and thus capture more high-order correlation between attributes in context.
	We consider that using an {\em order ensemble} is critical for pedestrian attribute modelling because:
	(1) Small sized training data makes poor model learning for
        most attribute classes; %renders the learned model not sufficiently accurate 
	%to predict most attribute classes;
	(2) Given significant attribute appearance change in
        surveillance data, the optimal sequential attribute
        correlation can vary significantly between different
        pedestrian images, with no single universally optimal sequential order. 
	Thus, we employ an ensemble of 10 attribute orders:
	rare first, frequent first, top-down and bottom-up 
	(for encoding body topological structure information), 
	global-local and local-global 
	(for interacting coarse and fine grained attributes),
	and 4 random orders
	(for incorporating randomness). % to attribute correlation mining).
	%
%	the some sequence prediction models in Computer Vision, such as image captioning, there is a natural order of the input sequence (e.g., words in a sentence). For some tasks when the order is not obvious, the order is pre-defined based on some heuristic rules. For instance, in the human pose estimation work of [9], a tree based ordering of joints is used.
%
%{\bf \color{blue} ** Address the comments **:\\
%	(1) how the order of attributes is designed \\
%	(2) predicting attributes sequentially is not natural \\

\vspace{0.1cm}
\noindent {\bf Model Training.}
For each attribute order in the ensemble,
we train an order-specific JRL model.
We learn any JRL model end-to-end by back-propagation through time \cite{werbos1990backpropagation}
so as to jointly optimise the encoder and decoder LSTM.
We use the cross-entropy loss on the softmax score
subject to training attribute labels. 
To %simplify the learning process
avoid noise back propagation 
from the RNN to CNN,
we do not train the CNN image feature representation network %(image feature extractor)
together with the JRL RNN encoder-decoder. % \cite{wang2016cnn,vinyals2015show}.
%
%\kiya{Also, we used 10 different attribute orders in ensemble: 6 random, plus body top-down, bottom-up,frequents first and rare first. (address Q4 from R1)} 
%
Each JRL model is optimised 
against per-image attribute sequences without duplication.
Therefore, repeated prediction is inherently penalised and discouraged.

%During training, each sub-model is optimised to {\em sequentially} predict
%attributes according to pre-defined order in a recurrent manner.
%Duplication in prediction is inherently penalised  
%with constraints built into the model optimisation because:
%(1) the model is trained without duplicated attributes on each
%training sample;
%(2) Predicting at step $t$ is conditioned both on
%explicitly on the attribute prediction of previous step $(t-1)$
%and implicitly on that from all previous steps
%given the recurrent feedback of LSTM.

\vspace{0.1cm}
\noindent {\bf Model Inference.}
Given a test image, each trained JRL model gives a multi-attribute prediction.
We generate a set of 10 predictions per test image given 10
order-specific models.
To infer the final prediction,
we adopt the majority voting scheme \cite{may1952set}.% by selecting the estimation receiving the maximal votes.
%% at least half (5) votes for a attribute class.
%(\sgg{why binary? i
%  thought that we are talking about multi-label attributes prediction
%  per image?})
% 
%In our evaluation, it is found that the ensemble of 
%JRL models 
%is effective to improve the sparsely labelled attribute recognition performance (Sec. \ref{sec:exp}).
%This makes sense due to the potential complementary
%effect between different recurrent correlations
%encoded in distinct attribute sequence orders.

%----------------------------------------------------------------------------
\section{Experiments}
\label{sec:exp}
\begin{figure}%[th!]
	\centering
	\includegraphics[width=1\linewidth]{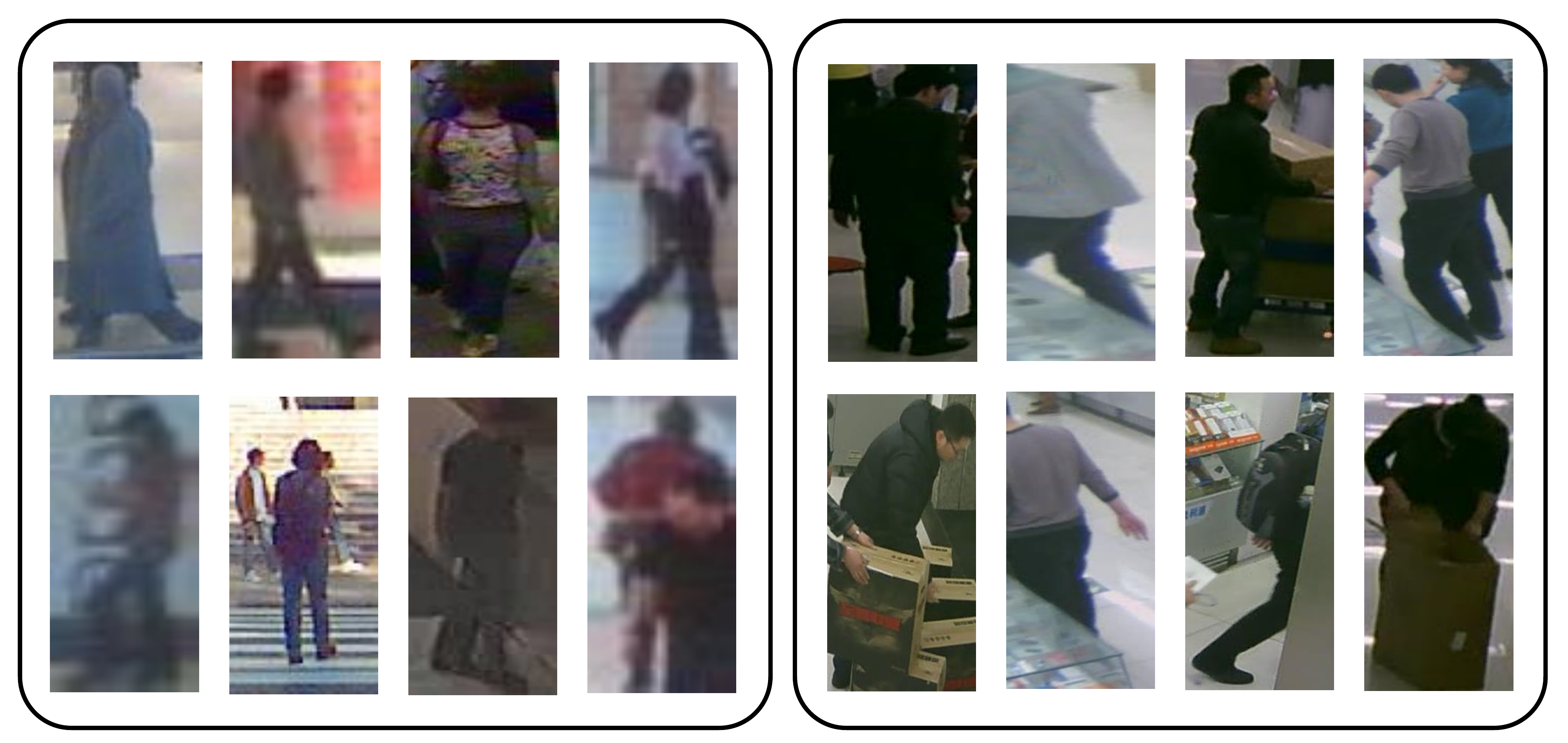}
	\vskip -0.1cm
	\caption{%\footnotesize
		Example images. {\bf Left}: %the two largest surveillance pedestrian
		%attribute benchmarks
                PETA~\cite{DENG2014PAR}; 
                {\bf Right}: RAP~\cite{li2016richly}.
		%		{\bf \color{cyan} TODO: Kiya to add more images with varying
		%		challenges such as occlusion, low-resolution, extreme lighting conditions.}
	}
	\label{fig:peta_rap}
	\vspace{-.1cm}
\end{figure}

%
%\subsection{Datasets and Settings}
%\label{sec:exp_setting}
% \textbf{PETA}~\cite{DENG2014PAR}: 
\noindent {\bf Datasets.}
For evaluations, we used the two largest pedestrian attribute datasets
publically available (Fig.\ref{fig:peta_rap}): 
{\bf (1)} The PEdesTrian Attribute ({\em PETA}) dataset \cite{DENG2014PAR}
consists of $19,000$ person images collected from $10$ small-scale
person datasets. Each image is labelled with 65 attributes (61 binary
+ 4 multi-valued). 
% $105$ attributes,
% 61 binary and 4 multi-valued attributes, 
%including demographics (e.g. gender and age range), 
%appearance (e.g. hair style), 
%upper and lower body clothing style (e.g. casual or formal), 
%and accessories.
%We also follows same parameter settings as \cite{DENG2014PAR}, randomly  divide dataset into three parts, 9500 for training, 1900 for verifying and 7600 for testing.
Following the same protocol per \cite{deng2015learning,li2015multi},
we randomly divided the whole dataset into three non-overlapping partitions: 
$9500$ for model training, $1900$ for verification, and $7600$ for model evaluation.
%Note that, the lack of a standard and consistent train/val/test data split
%can still affect the fairness in comparison
%and cause some inconsistency in results even using the exactly same model.
% 
{\bf (2)} The Richly Annotated Pedestrian ({\em RAP}) attribute
dataset \cite{li2016richly} has in total 41,585 images drawn from 26
indoor surveillance cameras. Each image is labelled with 72 attributes
(69 binary + 3 multiple valued).
% including XXX, 
% each image is also labelled with viewpoint, occlusion, and body parts.
We adopted the same data split as in \cite{li2016richly}:
33,268 images for training and the remaining 8,317 for test.
%The averaged results over 5 such random splits were used as the final performance.
We evaluated the same 51 binary attributes per \cite{li2016richly} for
a fair comparison.
%due to 
%the highly imbalanced attribute distribution.
%
%
% 
For both datasets, we converted multi-valued attributes into binary
attributes as in~\cite{deng2015learning,li2015multi,li2016richly}.
Both datasets pose significant challenges to pedestrian attribute
recognition under difficult illumination with low
resolution, occlusion and background clutter. 
%Table \ref{tab:dataset_compare} summarises the statistics of the two datasets.

%{\bf \color{blue} ** Address the comment **: \\
%	only one dataset is used for evaluation.}

%\begin{table} %[htbp]
%	\footnotesize
%	% \scriptsize
%	%\color{red}
%	\centering
%	\caption{
%		Statistics of PETA and RAP.
%	ID: Indoor; OD: Outdoor.}
%	\setlength{\tabcolsep}{0.08cm}
%	\begin{tabular}{c||c|c|c|c|c|c}
%		\hline
%		Dataset & Scene & Images & Attributes & Viewpoint & Occlusion & Part\\
%		\hline
%		PETA \cite{DENG2014PAR}  
%		& ID + OD  & 19,000 & 61 + 4  & No  & No & No \\
%		RAP \cite{li2016richly}   
%		& ID & 41,585 & 69 + 3 & Yes & Yes & Yes \\
%		\hline
%	\end{tabular}%
%	\label{tab:dataset_compare}%
%\end{table}%

%\begin{table}[htbp]
%%\footnotesize
%\scriptsize
%\color{red}
%  \centering
%  \caption{Comparison between two pedestrian attribute dataset}
%  \setlength{\tabcolsep}{0.01cm}
%    \begin{tabular}{rrrrrrrr}
%    \hline
%          & Camera & Scene & samples & Attributes& Viewpoint & Occlusion & Part location\\
%    \hline
%    PETA  & mix   & mix   & 19000 & 61    & no    & no    & no \\
%    RAP   & 26    & indoor & 41585 & 69    & 4 type & 4 type & 3 part \\
%    \hline
%    \end{tabular}%
%  \label{tab:dataset_compare}%
%\end{table}%

%========================
\vspace{0.1cm}
\noindent {\bf Performance Metrics.}
We use four metrics to evaluate attribute recognition performance.
{\bf (1)} Class-centric: For each attribute class, 
we compute the classification accuracy of positive and negative
samples respectively, average them to obtain an Average Precision
(AP) for this attribute, then take the mean of AP over all attributes
(mAP$^\text{cls}$) as the metric~\cite{deng2015learning}.
{\bf (2)} Instance-centric: We measure 
per instance (image) attribute prediction precision and recall. 
This measure additionally considers the inter-attribute correlation, 
in contrast to mAP that assumes independence between attributes \cite{li2016richly}. 
Specifically,
we compute the precision and recall of predicted attributes against
the groundtruth for each test image, and then 
take the mean of the two measures over all test images
to yield mean Precision (mPrc$^\text{ins}$) and mean Recall
(mRcl$^\text{ins}$) rates. 
We also compute the F1 score (F$1^\text{ins}$) 
based on mPrc$^\text{ins}$ and mRcl$^\text{ins}$
as a more comprehensive metric.

\vspace{0.1cm}
\noindent {\bf Competitors.}
%In our comparative evaluation, we considered a wide range of 
%alternative (a total $7$) state-of-the-art methods --
% \eddy{pedestrian attribute and multi-label image classification methods} .
We compared our model JRL against 7 contemporary and state-of-the-art
models. They include (I) two conventional discriminative attribute methods: 
{\bf (1)} We adopted CNN features
(FC$_7$ output of the AlexNet) with the SVM attribute model
\cite{layne2014attributes}, replacing its 
original Ensemble of Localized Features (ELF)
\cite{gray2008viewpoint,layne2014attributes};
{\bf (2)} MRFr2~\cite{deng2015learning} % \sgg
{is a graph based attribute
recognition method that exploits the context of neighbouring images by
Markov random field for mining the visual appearance proximity
relations between different images to support attribute reasoning};
(II) Three deep learning attribute recognition methods:
{\bf (3)} Attributes Convolutional Network
(ACN)~\cite{sudowe2015person} %\sgg
{trains jointly a CNN model for all
attributes, and sharing weights and transfer knowledge among different
attributes}; 
{\bf (4)} DeepSAR~\cite{li2015multi} %\sgg
{is a deep model that that treats
attribute classes individually by training  
multiple attribute-specific AlexNet models \cite{krizhevsky2012imagenet}};
{\bf (5)} DeepMAR~\cite{li2015multi}
%\sgg
{, unlike DeepSAR, considers
additionally inter-attribute correlation by jointly learning all
attributes in a single AlexNet model~\cite{krizhevsky2012imagenet}, so
to capture the concurrent attribute relationships, similar to
\cite{sudowe2015person}};
(III) One multi-person image annotation recurrent model: 
{\bf (6)} Contextual CNN-RNN (CTX CNN-RNN)~\cite{li2016sequential} %\sgg
{is
a CNN-RNN based sequential prediction model designed to encode the
scene context and inter-person social relations for modelling
multiple people in an image\footnote{In our weakly supervised setting for
attribute recognition, we have no attribute fine-grained location
labelling. So we feed the whole image CNN features at
each recurrent decoding step for both training and test.}};
(IV) One generic multi-label image classification model:
{\bf (7)} Semantically Regularised CNN-RNN (SR
CNN-RNN)~\cite{liu2016semantic} %\sgg
{is a state-of-the-art multi-label
image classification model that exploits the groundtruth attribute labels
for strongly supervised deep learning and richer image embedding}. 
%
%{\bf (7)} Recurrent Image Annotator (RIA) \cite{jin2016annotation}:
%Similar to SR CNN-RNN \cite{liu2016semantic}, 
%it is a strong CNN-RNN based multi-label image classification model 
%which utilises the transformed CNN image embedding as decoder input.  
%
%{\bf (8)} \cite{mun2016text}:
%

%
%
%Global neighbourhood constraint
%\cite{liang2016semantic}

%\begin{enumerate}
%	% \item ikSVM\cite{layne2014attributes}:
%	
%	% \item MRFr2\cite{DENG2014PAR}:
%	\item ACN \cite{sudowe2015person}:
%	\item DeepSAR \cite{li2015multi}:
%	\item DeepMAR \cite{li2015multi}:
%	\item SSDAL \cite{su2016deep}: 3 stage semi-supervised deep attribute learning framework
%	
%\end{enumerate}

\vspace{0.1cm}
\noindent \textbf{Implementation Details.}
%In our experiments, the CNN model (AlexNet \cite{krizhevsky2012imagenet})
%was firstly
%pre-trained on ILSVRC 2012 training dataset \cite{ILSVRC15} 
%and then fine-tuned on the PETA training data.
%Once trained,
%we utilised the last convolutional feature map of the CNN model for input image feature extraction.
The hidden state for both the encoder LSTM and the decoder LSTM of the
JRL model has $512$
units (neurons).
We set empirically the learning rate as $0.0001$ with AdamOptimizer~\cite{kingma2014adam},
%momentum rate $0.9$, 
%weight decay rate {\color{red} XXX}, 
and the dropout rate as $0.5$.
By default, we adopted the AlexNet (same as DeepMAR) as the network
architecture for image embedding, 
%unless stated otherwise, 
and top-2
exemplars were selected for inter-person context.

%{\bf \color{blue} ** Address this comment: **\\
%the large gap on reported and author implemented DeepMAR. 
%Overall improvement is limited, i.e. only 0.5\%
%}
 
%For fair comparison, we extract feature following \cite{li2015multi}, using CaffeNet structure pre-trained from imagenet and fine-tune on PETA. 
%We also follows same parameter settings as \cite{DENG2014PAR}, randomly  divide dataset into three parts, 9500 for training, 1900 for verifying and 7600 for testing.
\begin{table} [h]%[htbp]
	\centering 
	\footnotesize
	%\color{red}
	\setlength{\tabcolsep}{.2cm}
	\caption{Evaluation on PETA \cite{DENG2014PAR},
          $1^\text{st}$/$2^\text{nd}$ best results in red/blue.
		% *: results from \cite{li2016richly}
	}
	%**: results by our own implementation on our training/val/test data split. 
	\begin{tabular}{c|cccc}
		\hline
		\backslashbox{Method}{Metric}
		 & mAP$^\text{cls}$ %& Acc$^\text{ins}$ 
		& mPrc$^\text{ins}$ & mRcl$^\text{ins}$ & F$1^\text{ins}$ \\
		\hline \hline
		MRFr2\cite{deng2015learning}
		& 75.60 %& - 
		& - & - & - \\ 
		ELF+SVM \cite{layne2014attributes} 
		& 75.21 %& 43.68 
		& 49.45 & 74.24 
		& 59.36 \\
		CNN+SVM \cite{li2016richly}
		& 76.65 %& 45.41 
		& 51.33 & 75.14 & 61.00 \\
		\hline		
		ACN \cite{sudowe2015person} 
		& 81.15 %& 73.66 
		& \color{blue} 84.06 & 81.26 & 82.64 \\
		DeepSAR \cite{li2015multi} 
		& 81.30 %& - 
		& - & - & - \\
		% AlexNet + Softmax
		DeepMAR \cite{li2015multi} 
		& 82.60 %& 75.07 
		& 83.68 & \color{blue} 83.14 & \color{blue} 83.41 \\
		%		DeepMAR$^*$ \cite{li2015multi} 
		%		& 82.15 %& 75.07 
		%		& 83.25 & 82.70 & 82.97 \\
		\hline
		%		WPAL\_GoogleNet\_GMP \cite{yu2016weakly} 
		%		& 85.50 %& 75.07 
		%		& 84.07 & 85.78 & 84.90 \\		
		%		WPAL\_GoogleNet\_FSPP \cite{yu2016weakly} 
		%		& 84.16 %& 75.07 
		%		& 82.66 & 85.16 & 83.40 \\	
		
		%		GoogLeNet** \cite{szegedy2015going}
		%		& 72.56 %&  66.47 
		%		& 79.58 & 74.57 &76.99 \\
		%		GoogLeNet+{\bf JRL}**  
		%		&76.11 %& 70.46  
		%		& 79.67 &  80.51 &78.29  \\
		%		\hline
%		RIA \cite{jin2016annotation} 
%		& 83.45     % & {\bf 75.67}
%		&  83.96  & 83.97   & 83.89   \\
		
		CTX CNN-RNN \cite{li2016sequential} % AlexNet + JRL   %(global image ini/quick)
		& 80.13     % & {\bf 75.67}
		& 79.68   &  80.24  & 79.68   \\
		
		SR CNN-RNN \cite{liu2016semantic}
		& \color{blue} 82.83   % & {\bf 75.67}
		& 82.54 & 82.76 & 82.65  \\
		
		\hline
		% AlexNet + 
		\bf JRL
		& \color{red}  85.67    % & {\bf 75.67}
		& \color{red} 86.03 & \color{red} 85.34 & \color{red} 85.42  \\
		
		\hline
		%		GoogLeNet \cite{szegedy2015going}
		%        &80.47   %&  66.47 
		%		& 82.45  &81.28   &81.85  \\
		%		GoogLeNet + JRL   
		%		& 84.06 %& 70.46  
		%		&  84.97 & 84.78   & 84.76  \\
		%		\hline
	\end{tabular}%
	\label{tab:comp_arts_PETA}%
	\vspace{-0.1cm}
\end{table}%

\begin{table} [h]%[htbp]
	\centering 
	\footnotesize
	%\color{red}
	\setlength{\tabcolsep}{.2cm}
	\caption{Evaluation on RAP~\cite{li2016richly},
          $1^\text{st}$/$2^\text{nd}$ best results in red/blue.
	}
	\begin{tabular}{c|cccc}
		\hline
		\backslashbox{Method}{Metric} & mAP$^\text{cls}$ %& Acc$^\text{ins}$ 
		& mPrc$^\text{ins}$ & mRcl$^\text{ins}$ & F$1^\text{ins}$ \\
		\hline \hline
		MRFr2\cite{deng2015learning}
		& - %& - 
		& - & - & - \\ 
		ELF+SVM \cite{layne2014attributes} 
		& 69.94 %& 43.68 
		& 32.84 %& 74.24 
		& 71.18 & 44.95\\
		CNN+SVM \cite{li2016richly}
		& 72.28 %& 45.41 
		& 35.75 & 71.78 & 47.73 \\
		\hline		
		ACN \cite{sudowe2015person} 
		& 69.66 %& 73.66 
		& \color{red} 80.12 & 72.26 & \color{blue} 75.98 \\
		DeepSAR \cite{li2015multi} 
		& - %& - 
		& - & - & - \\
		DeepMAR \cite{li2015multi} 
		& 73.79 %& 75.07 
		& 74.92 & 76.21 & 75.56 \\
		\hline
		
%		RIA \cite{jin2016annotation}  
%		& 75.19 %& 75.07 
%		& 75.55 & 76.93 & 75.97 \\
		
		CTX CNN-RNN \cite{li2016sequential} % AlexNet + JRL   %(global image ini/quick)
		& 70.13 %& 75.07 
		& 71.03 & 71.20 & 70.23 \\
		
		SR CNN-RNN \cite{liu2016semantic}
		& \color{blue} 74.21 %& 75.07 
		& 75.11 & \color{blue} 76.52 & 75.83 \\
		
		\hline
		\bf JRL 
		& \color{red} 77.81  %& 75.07 
		& \color{blue} 78.11 & \color{red} 78.98  & \color{red} 78.58\\
		\hline
		
		%		GoogLeNet \cite{szegedy2015going}
		%		&  71.74   %&  66.47 
		%		&  73.27 & 73.01   & 73.26  \\
		%		GoogLeNet + JRL  
		%		& 75.68 %& 70.46  
		%		&  76.27 &75.97    &  75.94 \\
		%		\hline
	\end{tabular}%
	\label{tab:comp_arts_RAP}%
	\vspace{-0.2cm}
\end{table}%

%==============================================
\subsection{Comparison to the State-Of-The-Arts}
\label{sec:compare_arts}
Tables~\ref{tab:comp_arts_PETA} and \ref{tab:comp_arts_RAP} show
evaluations on PETA and RAP respectively.
It is evident that the proposed JRL model achieves the best accuracy
on PETA given by all four evaluation metrics, and on RAP the best
accuracy given by three metrics except mPrc$^\text{ins}$ coming second
best (JRL 78.11\% vs. ACN 80.12\%). This implies that ACN is
conservative in prediction, i.e. predicting only very confident
attributes. More significantly, JRL outperforms in mAP$^\text{cls}$ the
state-of-the-art attribute model DeepMAR~\cite{li2015multi} and
multi-label image annotation model SR CNN-RNN~\cite{liu2016semantic}
respectively by $3.07\%$ and $2.84\%$ on PETA,
$4.02\%$ and $3.60\%$ on RAP.
{Similar margins are observed with other performance metrics,
except with the mPrc$^\text{ins}$ ACN \cite{sudowe2015person} achieves
the best score ($80.12\%$ vs. $78.11\%$ by JRL) on RAP,
but with a much lower mRcl$^\text{ins}$ ($72.26\%$ vs. $78.98\%$ by JRL)
also a lower overall F1$^\text{ins}$ ($75.98\%$ vs. $78.58\%$ by JRL).}
This shows clearly the benefit of the proposed correlation and
context joint recurrent learning approach to predicting ambiguous
pedestrian attributes in poor quality surveillance images. 
%, as compared to either non-deep methods 
%such as SVM and MRFr2 or deep alternatives like the state-of-the-art DeepMAR and ACN models.
This is mainly due to JRL's capacity to maximise and exploit
the complementary effect of correlation and context on sparsely
labelled weak annotations, through an end-to-end encoder/decoder learning.

\begin{figure}%[th!]
	\centering
	\includegraphics[width=1\linewidth]{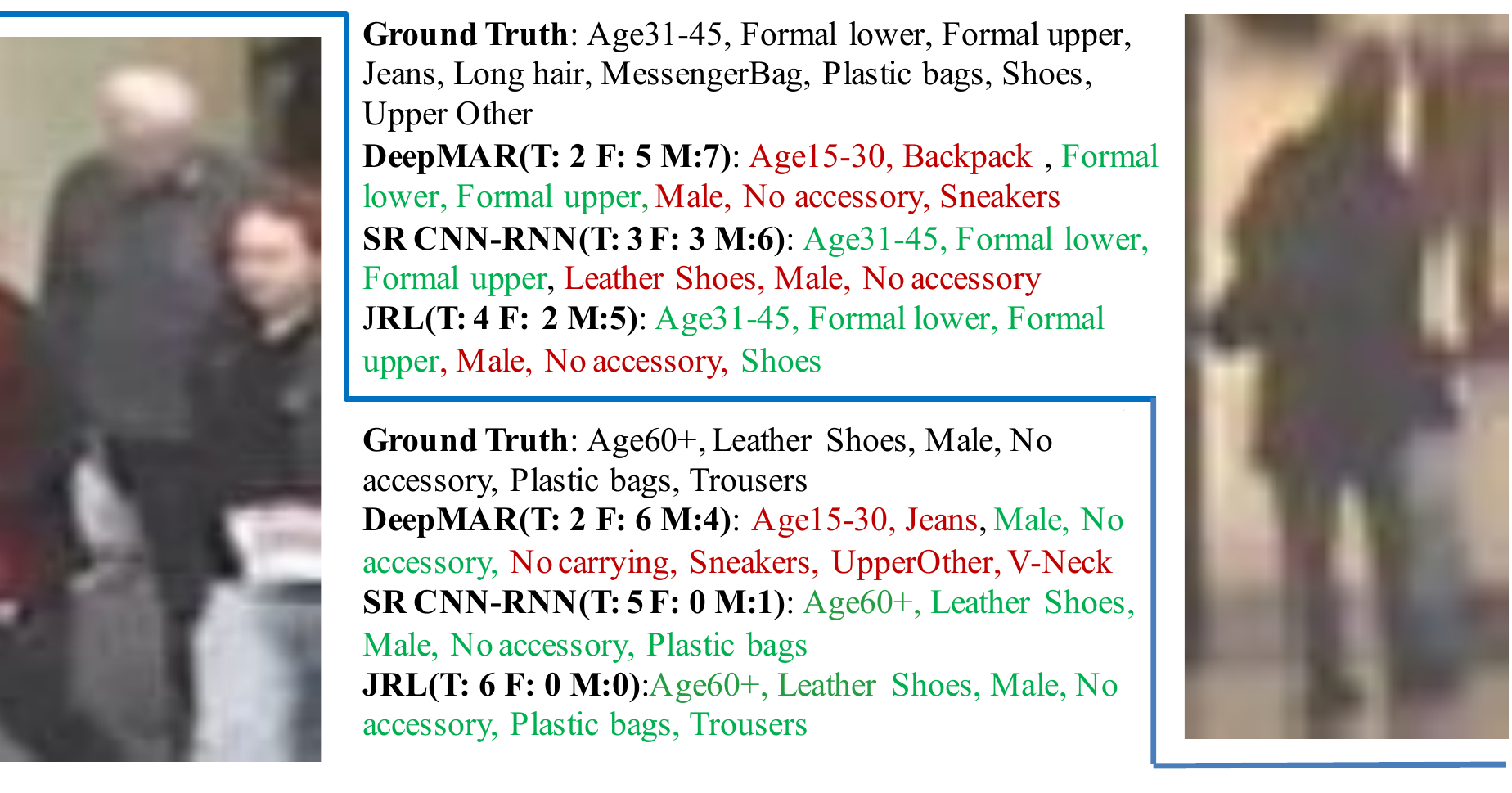}
	\vskip -0.2cm
	\caption{%\footnotesize
		Qualitative evaluation of pedestrian attribute recognition
		on PETA~\cite{DENG2014PAR}, with wrong predictions in
                red, true in green.
		%{\bf \color{cyan} TODO: Kiya to do 
		%	(1) Based on Table 4, select 5 attributes with different characteristics 
		%	such as small/large, subtle to recognise, easy to be occluded;
		%	(2) Based on Table 2/3, select top-3 methods;
		%	(3) Select 6 images as example for each dataset: covering easy, middle, and hard cases,
		%	with different challenge factors, such as occlusion, background clutters. 
		%	Our JRL model is not necessarily perfect for every example, since our method 
		%	is clearly not perfect from Table 2/3.}
		%	\color{blue}{}
	}
	\label{fig:example}
	\vspace{-.1cm}
\end{figure}
% SGG: for journal extension version - question: why these particular
% 35 classes?
%\sgg{To further examine the model performance, %on individual attributes, 
%	we compare the AP results on $35$ PETA attributes in Table \ref{tab:35att}.
%	%We have several observations as below:
%	It is found that the JRL achieves the best accuracies 
%	on 26 attributes, % out of 35 attributes,
%	but performs not well on a few such as
%	``Sandals'', ``SunGlasses'', and ``V-Neck''.
%	These attributes rarely appear and intuitively not highly correlated
%	with others. 
%	It is therefore not surprised that the best model 
%	for them is the DeepSAR model which considers no latent correlation.
%	This suggests the weak dependency between these attributes with
%	others, and also
%	justifies the low accuracies of all sequential
%	methods include the JRL.
%        }

%%==============================================
%\subsection{Robustness Against Training Label Sparsity}
%\label{sec:eval_sparcity}
\noindent {\bf Robustness Against Training Label Sparsity.}
In addition to the overall performance comparisons, we further
conducted a model scalability evaluation against the training data
size to better understand model robustness to small sized
(difficult-to-collect) attribute labels.
%due to high annotation cost. 
To that end, we randomly removed varying ratios
($25\!\!\sim\!\!75\%$) of the whole training data set
with the test data remaining unchanged on both PETA and RAP.
We compared the JRL model with 
the best two competitors: DeepMAR \cite{li2015multi} and SR CNN-RNN \cite{liu2016semantic}. 
%This evaluation allows to further demonstrate the robustness of 
%distinct methods against varying numbers of training data.
%We show the results in Table \ref{tab:sparcity}.
It is evident from Table~\ref{tab:sparcity} 
that JRL is more robust than
both DeepMAR and SR CNN-RNN against training data sparsity.
When training data decreased from $100\%$ to $25\%$,
the mAP$^\text{cls}$ performance drop of JRL is $3.64\%$ and $3.55\%$
on PETA and RAP respectively. This compares favourably against DeepMAR ($6.23\%$ and
$5.73\%$) and SR CNN-RNN ($6.24\%$ and $5.85\%$).
%Similar trends are found on other metrics.
%Our JRL model gain even larger both absolute and relative performance improvement over DeepMAR and SR CNN-RNN
%when sparser training data are available.
%\sgg
{This validate the advantages and potentials of our proposed JRL model
in handling sparse training data situations by 
effectively maximising the joint benefits of 
modelling attribute context and correlation end-to-end
from only limited labelled data.}
%{\bf \color{cyan} TODO: need to cover RAP when results are ready.}
%e.g., the performance drop of JRL is given less amount of labels (from 100\% to 25\% training data size), 
%the performance drops of our JRL model is smaller than the most competitive DeepMAR**, 3.10 compare to 5.49 on mAP, and other metric witness same trends as well. 

\begin{table} [h]%[htbp]
	\centering 
	\footnotesize
	%\color{red}
	\setlength{\tabcolsep}{.05cm}
	\caption{
		Model robustness vs. training data size
                (TDS) in \%.
		% **: results by our own implementation on our train/val/test data splits. 
%		{\bf {\color{cyan} TODO for Kiya: do experiments on PEAT and RAP. 
%				%We need to add one more method, based on their performance in Table 2/3.
%			}}
	}
	\begin{tabular}{c|c|c|cccc}
		\hline
		Dataset
		 & TDS (\%) &  \backslashbox{Model}{Metric} 
		& mAP$^\text{cls}$ %& Acc$^\text{ins}$ 
		& mPrc$^\text{ins}$ & mRcl$^\text{ins}$ & F$1^\text{ins}$ \\
		\hline \hline
		\multirow{12}{*}{PETA \cite{deng2015learning}}
		& \multirow{3}{*}{100} & DeepMAR \cite{li2015multi} 
		& 82.60 %& 75.07 
		& 83.68 & 83.14 & 83.41 \\
		\cline{3-7}
		& & SR CNN-RNN\cite{liu2016semantic}
		&  82.83    % & {\bf 75.67}
		&  82.54 & 82.76 &  82.65\\
		\cline{3-7}
		& & {\bf JRL} 
		& \bf 85.67    % & {\bf 75.67}
		& \bf 86.03 & \bf 85.34 & \bf 85.42 \\
		\cline{2-7}
		& \multirow{3}{*}{75} & DeepMAR \cite{li2015multi} 
		& 80.83 % & 71.77 
		& 81.02 & 81.73 & 81.37
		\\
		\cline{3-7}
		& & SR CNN-RNN\cite{liu2016semantic}
		&  81.06    % & {\bf 75.67}
		&  81.11 & 81.66 &  81.21\\
		\cline{3-7}
		& & {\bf JRL} 
		& {\bf 84.45} %& {\bf 74.47 }
		& {\bf 84.86} & {\bf 84.23} & {\bf 84.07 }
		\\
		\cline{2-7}
		& \multirow{3}{*}{50} & DeepMAR \cite{li2015multi} 
		& 79.16 %& 69.13 
		& 80.66 & 80.39& 80.52  
		\\
		\cline{3-7}
		& & SR CNN-RNN\cite{liu2016semantic}
		&  79.09   % & {\bf 75.67}
		&  80.40 & 80.13 &  80.06\\
		\cline{3-7}
		& & {\bf JRL} 
		& {\bf 83.42} %& {\bf 73.69 }
		& {\bf 84.16} & {\bf 82.39} & {\bf 82.46}
		\\
		\cline{2-7}
		& \multirow{3}{*}{25} & DeepMAR \cite{li2015multi} 
		& 76.37 % & 67.25 
		& 79.12 &77.93 & 78.52  
		\\
		\cline{3-7}
		& & SR CNN-RNN\cite{liu2016semantic}
		&  76.59   % & {\bf 75.67}
		&  79.23 & 78.12 &  78.39\\
		\cline{3-7}
		& & {\bf JRL}
		& {\bf 82.03 } %& {\bf 71.34 }
		& {\bf 83.16} & {\bf 81.01} & {\bf 81.51 }
		\\
		\hline\hline
		\multirow{12}{*}{RAP \cite{li2016richly}}
		& \multirow{3}{*}{100} & DeepMAR \cite{li2015multi} 
		& 73.79 %& 75.07 
		& 74.92 & 76.21 & 75.56  \\
		\cline{3-7}
		& & SR CNN-RNN\cite{liu2016semantic}
		&  74.21   % & {\bf 75.67}
		&  75.11 & 76.52 &  75.83\\
		\cline{3-7}
		& & {\bf JRL} 
		& \bf 77.81  %& 75.07 
		& \bf 78.11 & \bf 78.98  & \bf 78.58 \\
		\cline{2-7}
		& \multirow{3}{*}{75} & DeepMAR \cite{li2015multi} 
		& 71.38 % & 67.25 
		& 72.40 &74.62 & 73.49 \\
		\cline{3-7}
		& & SR CNN-RNN\cite{liu2016semantic}
		&  71.96   % & {\bf 75.67}
		&  72.33 & 74.73 &  73.64\\
		\cline{3-7}
		& & {\bf JRL} 
		& \bf 76.69  %& 75.07 
		& \bf 77.34 & \bf 77.76  & \bf 77.36 \\
		\cline{2-7}
		& \multirow{3}{*}{50} & DeepMAR \cite{li2015multi} 
		& 70.01 % & 67.25 
		& 71.52 &72.53 & 72.06 \\
		\cline{3-7}
		& & SR CNN-RNN\cite{liu2016semantic}
		&  70.53   % & {\bf 75.67}
		&  71.96 & 72.77 &  72.36\\
		\cline{3-7}
		& & {\bf JRL} 
		& \bf 75.51  %& 75.07 
		& \bf 76.31 & \bf 76.69  & \bf 76.64 \\
		\cline{2-7}
		& \multirow{3}{*}{25} & DeepMAR \cite{li2015multi} 
		& 68.06 % & 67.25 
		& 70.33 &69.86 & 70.08 \\
		\cline{3-7}
		& & SR CNN-RNN\cite{liu2016semantic}
		&  68.36   % & {\bf 75.67}
		&  70.67 & 70.39 &  70.67\\
		\cline{3-7}
		& & {\bf JRL}
		& \bf 74.26  %& 75.07 
		& \bf 75.16 & \bf 75.21  & \bf 75.34 \\
		\hline
	\end{tabular}%
	\label{tab:sparcity}%
\end{table}%

%========================================
\subsection{Further Analysis and Discussions}
\label{sec:eval_JRL}
%\eddy{We further evaluated and analysed the proposed JRL approach in these aspects: 
%{\bf (1)} Effects of intra-person topological context; 
%{\bf (2)} Effects of inter-person compensation context;
%{\bf (3)} Effects of model ensemble;
%{\bf (4)} Effects of localised image region number;
%{\bf (5)} Effects of recurrent attribute attention;
%{\bf (6)} Generalisation with different CNN models;
%{\bf (7)} Fusion analysis of inter-person compensation context;
%{\bf (8)} Examination of discovered attribute correlations by the JRL model.
%}

\vspace{0.1cm}
\noindent {\bf (1) Benefit of intra-person attribute context. }
We evaluated explicitly the benefit of modelling intra-person attribute context 
by the encoder LSTM.
For that, we built a stripped-down JRL model by
removing the encoder and directly using the CNN FC features for the
decoder input.
% We call this model as JRL (No LPSC).
% the performance between our encoder+decoder and
%our decoder alone that takes as input the image level features 
%(e.g., from FC$_1$ activations of AlexNet), instead of 
%the context vector from the encoder.
% 
Table \ref{tab:eval_local_context} shows the difference on both PETA
and RAP, improving mAP$^\text{cls}$ by $2.22\%$ and $2.62\%$
respectively, similarly under the other metrics.
%
%We reported the evaluations in Table \ref{tab:eval_context}.
%It is shown that with context correlation learning,
%we can obtain some additional information, although not significant.
%This plausible reason may be due to the large noise
%caused from complex/cluttered background and 
%challenging viewing conditions.

%{\bf \color{blue} ** A comment to consider and address **: \\ 
%	what will happen if the 6 strips are fed randomly.
%	(We may consider a reverse order, i.e. bottom-up, and see 
%	what can be obtained?)
%}

\begin{table}[h]
	\centering 
	\footnotesize
	%\color{red}
	\setlength{\tabcolsep}{.15cm}
	\caption{Benefit of intra-person Attribute Context (AC).
		%{\bf \color{cyan} TODO: Kiya to do}
	}
	\begin{tabular}{c|c||cccc}
		\hline
		
		Dataset & \backslashbox{Method}{Metric} & mAP$^\text{cls}$ %& Acc$^\text{ins}$ 
		& mPrc$^\text{ins}$ & mRcl$^\text{ins}$ & F$1^\text{ins}$ \\
		\hline \hline
		\multirow{2}{*}{PETA \cite{deng2015learning}} & 
		JRL (No AC) 
		&  83.45    % & {\bf 75.67}
		& 83.96 &  83.97 &  83.89 \\
		& JRL  
		& \bf 85.67    % & {\bf 75.67}
		& \bf 86.03 & \bf 85.34 & \bf 85.42 \\	
		\hline
		\multirow{2}{*}{RAP \cite{li2016richly}} & 
		JRL (No AC) 
		& 75.19  %& 75.07 
		&  75.55 &  76.93 & 75.97 \\
		& JRL 
		& \bf 77.81  %& 75.07 
		& \bf 78.11 & \bf 78.98  & \bf 78.58 \\
		\hline
	\end{tabular}%
	\label{tab:eval_local_context}%
	\vspace{-0.3cm}
\end{table}%

\vspace{0.1cm}
\noindent {\bf (2) Effect of inter-person similarity context.}
We also evaluated explicitly the benefit of exploiting auxiliary
exemplar images as inter-person context. For that,
we excluded them in both model training and inference stages.
Table \ref{tab:eval_exemplar_context} shows that 
this context modelling brings $0.65\%$ and $0.87\%$ boost in
mAP$^\text{cls}$ on PETA and RAP respectively.

\begin{table}[h]
	\centering 
	\footnotesize
	%\color{red}
	\setlength{\tabcolsep}{.15cm}
	\caption{
		Effect of inter-person Similarity Context (SC).
		%{\bf \color{cyan} TODO: Kiya to do missing experiments}
	}
	\begin{tabular}{c|c||cccc}
		\hline
		Dataset & \backslashbox{Method}{Metric} & mAP$^\text{cls}$ %& Acc$^\text{ins}$ 
		& mPrc$^\text{ins}$ & mRcl$^\text{ins}$ & F$1^\text{ins}$ \\
		\hline \hline
		\multirow{2}{*}{PETA \cite{deng2015learning}} & 
		JRL(No SC) 
		& 85.02    % & {\bf 75.67}
		& 85.27 &  84.36 &  84.86 \\
		& JRL 
		& \bf 85.67    % & {\bf 75.67}
		& \bf 86.03 & \bf 85.34 & \bf 85.42 \\
		\hline
		\multirow{2}{*}{RAP \cite{li2016richly}} & 
		JRL(No SC) 
		&  76.94  %& 75.07 
		&  77.39 &  78.23  & 77.92 \\
		& JRL
		& \bf 77.81  %& 75.07 
		& \bf 78.11 & \bf 78.98  & \bf 78.58 \\
		\hline
	\end{tabular}%
	\label{tab:eval_exemplar_context}%
	\vspace{-0.3cm}
\end{table}%

\vspace{0.1cm}
\noindent {\bf (3) Effects of model ensemble.}
We evaluated the benefit of exploiting attribute order ensemble.
%	 on ambiguous 
%	pedestrian attribute modelling.
	We compared the average results of all 10 orders.
	Table~\ref{tab:eval_ensemble} shows that 
	the ensemble of {\em distinct sequential orders} 
	improves significantly the model performance, improving
        mAP$^\text{cls}$ by $3.54\%$ and $3.07\%$ on PETA and RAP when
        compared to the average. 
	{This validates our ensemble design intuition
	for modelling ambiguous attributes in poor surveillance
	images from sparsely labelled training data.}
%	the ensemble of high-order attribute correlation learned
%	from distinct orders is able to effectively improve the model performance,
%	with the average mAP$^\text{cls}$ $3.54\%$(85.67-82.13)/$XX\%$ lower than
%	the ensemble result, 
%	and the best single-order mAP$^\text{cls}$ $0.86\%$(85.67-84.81)/$XX\%$
%	on PETA and RAP, respectively.
%	Specifically, we have these observations:
%	(1) On PETA, ``Rare First'' order yields the best result whilst ``Frequent First'' is the poorest order. 
%	This shall be due to collectively 
%	the limited training data size ($9,500$ images) and 
%	attribute recognition challenges in surveillance images,
%	More specifically, it is difficult for the model to produce
%	sufficiently reliable predictions of most frequent attributes 
%	(e.g. LowerBodyCasual, UpperBodyCasual) to help the reasoning
%	of less frequent ones. 
%	In the meanwhile, rare attributes (e.g. Logo, V-Neck) gain no promotion and order priority in model optimisation. 
%	(2) ``Global-Local'' order is better than ``Local-Global''.
%	This may be explained by XXXX the higher difficulty of 
%	recognising local attributes (e.g. XXXX)
%	(3) Both ``Top-Down'' and ``Bottom-Up'' orders work similarly well.
%	This suggests the goodness of regular attribute spatial orders,
%	which is intuitive in general.
%	(4) ``Random'' orders perform reasonably well but mostly worse
%	than statistics and topology based ones except ``Frequent First''.
%
%{\bf \color{blue} ** Address this comment **: \\
%	how much does the ensemble contribute.
%}

\begin{table}[h]
	\centering 
	\footnotesize
	%\color{red}
	\setlength{\tabcolsep}{.15cm}
	\caption{Effects of the model ensemble.
		%{\bf \color{cyan} TODO for Kiya}
	}
	\begin{tabular}{c|c||cccc}
		\hline
		Dataset & \backslashbox{Method}{Metric} & mAP$^\text{cls}$ %& Acc$^\text{ins}$ 
		& mPrc$^\text{ins}$ & mRcl$^\text{ins}$ & F$1^\text{ins}$ \\
		\hline \hline
		\multirow{2}{*}{PETA \cite{deng2015learning}} & 
		\em Average 
		&  82.13    % & {\bf 75.67}
		&  82.55 & 82.12 & 82.02 \\
		\cline{2-6}
		& \bf Ensemble
		& \bf 85.67    % & {\bf 75.67}
		& \bf 86.03 & \bf 85.34 & \bf 85.42 \\
		\hline %\hline
		\multirow{2}{*}{RAP \cite{li2016richly}}
		& \em Average
		&74.74  & 75.08  & 74.96  & 74.62  \\
		\cline{2-6}
		& \bf Ensemble
		& \bf 77.81  %& 75.07 
		& \bf 78.11 & \bf 78.98  & \bf 78.58 \\
		\hline
	\end{tabular}%
	\label{tab:eval_ensemble}%
	\vspace{-0.3cm}
\end{table}%

\vspace{0.1cm}
\noindent {\bf (4) Effects of recurrent attribute attention.}
Table~\ref{tab:eval_attention} shows that the data-driven alignment
between image region sequence and attribute label sequence
is beneficial, giving $1.64\%$ and $1.85\%$ in mAP$^\text{cls}$ boost
on PETA and RAP.
\begin{table}[h]
	\centering 
	\footnotesize
	%\color{red}
	\setlength{\tabcolsep}{.15cm}
	\caption{Effects of recurrent attribute attention.
%		(\sgg{need to keep it as it's part of our
%            contributions in the intro})
		%{\bf \color{cyan} TODO: Kiya to do missing experiments}
	}
	\begin{tabular}{c|c||cccc}
		\hline
		Dataset & \backslashbox{Method}{Metric}  & mAP$^\text{cls}$ %& Acc$^\text{ins}$ 
		& mPrc$^\text{ins}$ & mRcl$^\text{ins}$ & F$1^\text{ins}$ \\
		\hline \hline
		\multirow{2}{*}{PETA \cite{deng2015learning}} & 
		JRL(No Attention) 
		& 84.03    % & {\bf 75.67}
		& 84.92 &  84.19 &  84.24 \\
		& JRL 
		& \bf 85.67    % & {\bf 75.67}
		& \bf 86.03 & \bf 85.34 & \bf 85.42 \\
		\hline
		\multirow{2}{*}{RAP \cite{li2016richly}} & 
		JRL(No Attention) 
		& 75.96  %& 75.07 
		&  76.89 &  77.49  & 77.13 \\
		& JRL
		& \bf 77.81  %& 75.07 
		& \bf 78.11 & \bf 78.98  & \bf 78.58 \\
		\hline
	\end{tabular}%
	\label{tab:eval_attention}%
\end{table}%

\begin{figure}%[th!]
	\centering
	\includegraphics[width=1\linewidth]{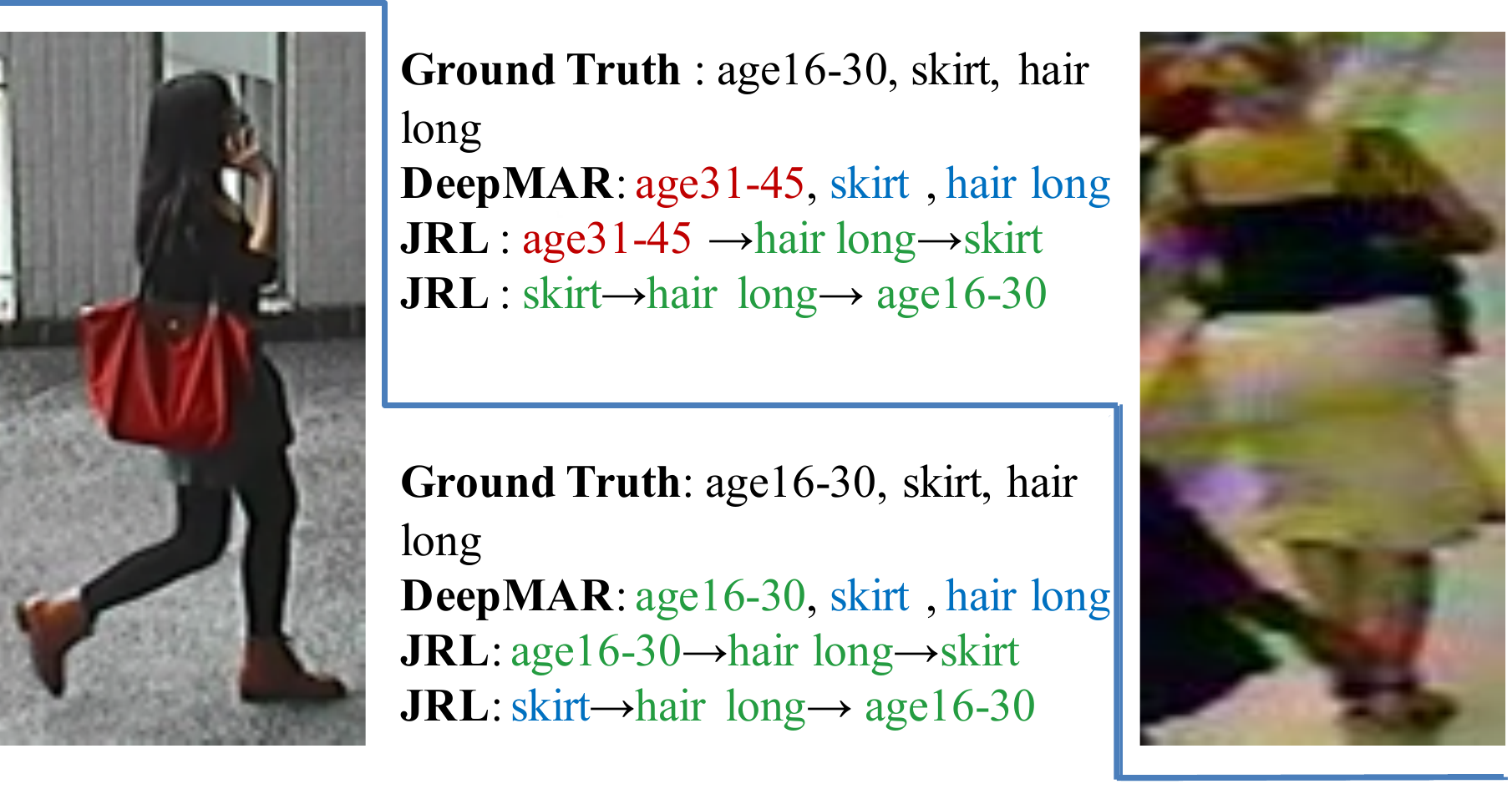}
	\vskip -0.1cm
	\caption{%\footnotesize
		Qualitative analysis of latent attribute correlation, with
		wrong predictions in red, true in green and missed
                predictions in blue.
        The two examples are from PETA \cite{DENG2014PAR}.        
		%{\bf \color{cyan} TODO: Kiya to make a figure.}
	}
	\label{fig:example_correlation}
	\vspace{-.2cm}
\end{figure}

\vspace{0.1cm}
\noindent {\bf (5) Qualitative analysis on the effect of attribute correlations.}
We examined more carefully the effect of attribute correlations on
the JRL model performance.
Fig.~\ref{fig:example_correlation} shows two examples. The person on
the left with a
fashionable bag and legging/shoes was predicted reliably by JRL to be
``young'', whilst the ``hair long'' is less obvious and ``skirt''
almost invisible but both predicted correctly by JRL invoking
the relevant ordering context.
In contrast, a non-sequence prediction model DeepMAR
\cite{li2015multi} missed both ``hair long'' and
``skirt''. This is because that JRL benefited from identifying the
relevant sequential ``age-hair-skirt'' ordering as inter-attribute
correlation context for attribute prediction. When getting the wrong
ordering context, JRL missed ``skirt''. The person on the right
wears ``skirt'', clearly visible, with both ``hair long'' and ``age''
unclear. The JRL model 
again benefited from invoking the useful ``skirt-hair-age'' ordering
context for attribute prediction. When given the wrong ordering, JRL
makes the mistake on ``age'' prediction. DeepMAR missed both
``skirt'' and ``hiar long'', and predicted ``age'' incorrectly.
	%{\bf \color{cyan} TODO: Kiya to do the followings.
	%Identify some attribute pairs in the model ensemble, under some order, the attributes can be predicted more accurately. Also, show some visual examples for them. \\
%	(\sgg{keep this - this is a very good analysis.})

%----------------------------------------------------------------------------
\section{Conclusion}

In this work, we presented a novel deep Joint Recurrent Learning (JRL)
model for exploring attribute context and correlation in deep learning
of pedestrian attributes given low quality surveillance images and
small sized training data.
Our JRL method outperforms a wide range of state-of-the-art 
pedestrian attribute and multi-label classification methods. Extensive
experiments demonstrate the
advantages and superiority of joint learning high-order (sequential)
inter-attribute correlation on two pedestrian attribute benchmarks. 
Moreover, the JRL model is shown to be more robust than
state-of-the-art deep models when trained with small sized training
data, thus more scalable to real-world applications with
limited annotation budget available.

\section*{Acknowledgements}
%\vspace{-0.2cm}
\noindent This work was partially supported by the China Scholarship Council, Vision Semantics Ltd., and Royal Society Newton Advanced Fellowship Programme (NA150459).

{\small
\bibliographystyle{ieee}
\bibliography{ped_attr}
}

\end{document}